\documentclass[10pt,twocolumn,letterpaper]{article}

\usepackage{iccv}
\usepackage{times}
\usepackage{epsfig}
\usepackage{graphicx}
\usepackage{amsmath}
\usepackage{amssymb}
\usepackage{capt-of}
\usepackage{multirow}
\usepackage{makecell}
\usepackage{booktabs}
\usepackage{amsmath}
\makeatletter
\@namedef{ver@everyshi.sty}{}
\makeatother
\usepackage{tikz}
\usepackage{array}
\usepackage{color, colortbl}

\usepackage[pagebackref=true,breaklinks=true,colorlinks,bookmarks=false]{hyperref}

\iccvfinalcopy %

\def\iccvPaperID{12354} %

\ificcvfinal\fi

\begin{document}

\title{PRedItOR: Text Guided Image Editing with Diffusion Prior}
\author{Hareesh Ravi
\and
Sachin Kelkar
\and
Midhun Harikumar
\and 
Ajinkya Kale \and \\ 
Adobe Applied Research}

\twocolumn[{%
\renewcommand\twocolumn[1][]{#1}%
\maketitle
\vspace*{-0.5cm}
\begin{center}
 \begin{tabular}{c@{\hspace{0.5cm}}c@{\hspace{0.1cm}}c@{\hspace{0.1cm}}c@{\hspace{0.1cm}}c@{\hspace{0.1cm}}c@{\hspace{0.1cm}}}
 \centering
 base image&\emph{a bluejay}&\emph{a crow}&\emph{an owl}&\emph{a parrot}&\emph{northern cardinal}\\
 \includegraphics[width=.15\linewidth]{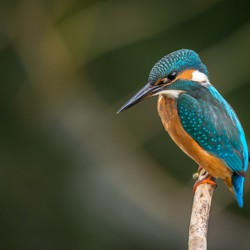} \hspace{0.1in}&
 \includegraphics[width=.15\linewidth]{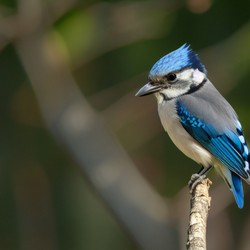}&
 \includegraphics[width=.15\linewidth]{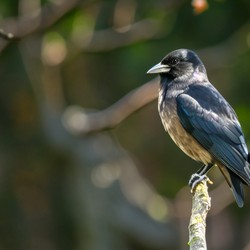}&
 \includegraphics[width=.15\linewidth]{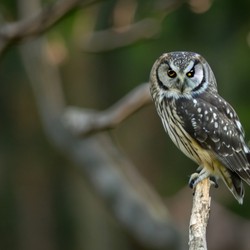}&
 \includegraphics[width=.15\linewidth]{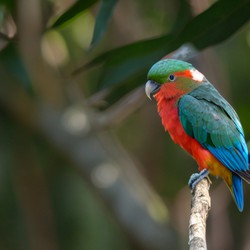}&
 \includegraphics[width=.15\linewidth]{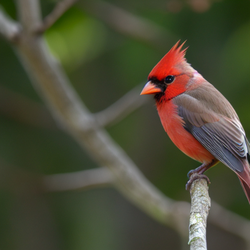} \\
 &\emph{dog}&\emph{watercolor painting}&\emph{pencil sketch}&\emph{terminator robot}&\emph{plush toy}\\
 \includegraphics[width=.15\linewidth]{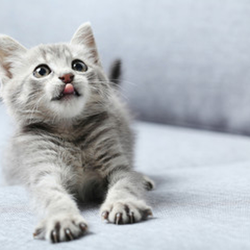} \hspace{0.1in}&
 \includegraphics[width=.15\linewidth]{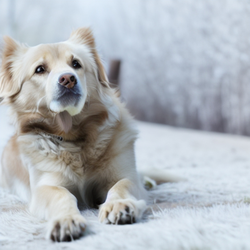}&
 \includegraphics[width=.15\linewidth]{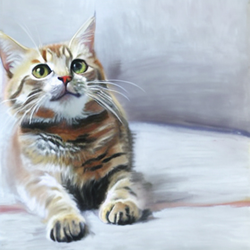}&
 \includegraphics[width=.15\linewidth]{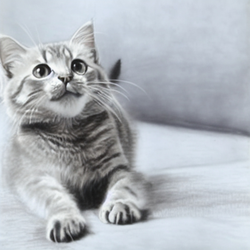}&
 \includegraphics[width=.15\linewidth]{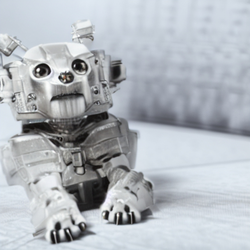}&
 \includegraphics[width=.15\linewidth]{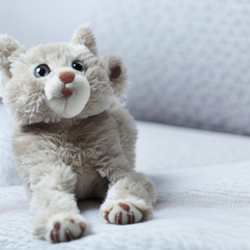} \\
 &\emph{black dog in flood}&\emph{back dog in nyc}&\emph{black dog in fall}&\emph{black dog on beach}&\emph{black dog on boat}\\
 \includegraphics[width=.15\linewidth]{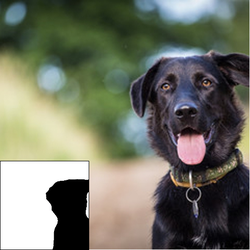} \hspace{0.1in}&
 \includegraphics[width=.15\linewidth]{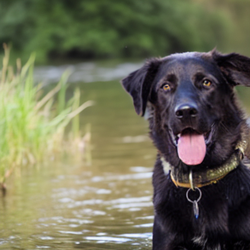}&
 \includegraphics[width=.15\linewidth]{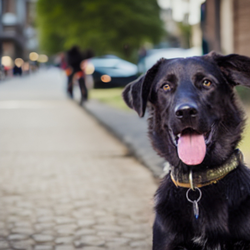}&
 \includegraphics[width=.15\linewidth]{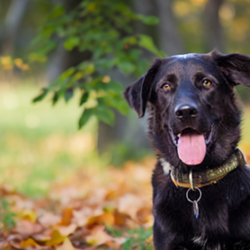}&
 \includegraphics[width=.15\linewidth]{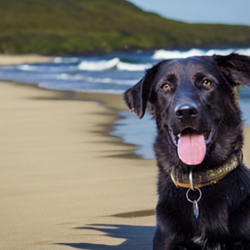}&
 \includegraphics[width=.15\linewidth]{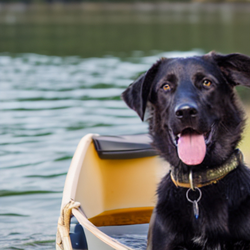} \\
 \end{tabular}
 \captionof{figure}{Examples of text guided image editing using PRedItOR. PRedItOR uses a Diffusion Prior model to perform conceptual editing in the CLIP image embedding space. The generated edit embedding is used to perform a final structural edit step to complete the edit. Both steps have a weight parameter to control the strengths. Our method does not require base prompts, optimization of embeddings, finetuning of model weights, training models or additional objectives. The generated edits are structure preserving and can be enhanced by providing an optional mask (last row: mask provided along with base image; white indicates region to be edited).}
 \label{fig:intro}
\end{center}}]

\maketitle
\ificcvfinal\thispagestyle{empty}\fi

\begin{abstract}
\vspace{-0.1in}
Diffusion models have shown remarkable capabilities in generating high quality and creative images conditioned on text. An interesting application of such models is structure preserving text guided image editing. Existing approaches rely on text conditioned diffusion models such as Stable Diffusion or Imagen and require compute intensive optimization of text embeddings or fine-tuning the model weights for text guided image editing. We explore text guided image editing with a Hybrid Diffusion Model (HDM) architecture similar to DALLE-2. Our architecture consists of a diffusion prior model that generates CLIP image embedding conditioned on a text prompt and a custom Latent Diffusion Model trained to generate images conditioned on CLIP image embedding. We discover that the diffusion prior model can be used to perform text guided conceptual edits on the CLIP image embedding space without any finetuning or optimization. We combine this with structure preserving edits on the image decoder using existing approaches such as reverse DDIM to perform text guided image editing. Our approach, PRedItOR does not require additional inputs, fine-tuning, optimization or objectives and shows on par or better results than baselines qualitatively and quantitatively. We provide further analysis and understanding of the diffusion prior model and believe this opens up new possibilities in diffusion models research.

\end{abstract}

\section{Introduction}
\label{sec:intro}
Diffusion Models (DM) have shown great capabilities in generating high quality, diverse images conditioned on text \cite{ddpm, diffusion_beat_gans}. Large scale training of diffusion models further improved the quality and relevance of generations \cite{dalle2, imagen, ldm}. Since then, many approaches have been proposed to improve their training and sampling speed \cite{ldm, ddim, dpm_solver_pp}, controllability \cite{diffusion_beat_gans, classifier_free_guidance}, and quality \cite{ediff_i, imagen, dalle2} and to enable downstream applications \cite{sdedit, diffae, ediff_i, make_a_video, imagen-video} such as editing, style transfer, video generation etc. One important application that garnered a lot of attention is text guided image editing. Many approaches \cite{prompt2prompt, imagic, unitune, plugnplay, shape} have been proposed to leverage the pretrained knowledge of large text-to-image diffusion models for text guided image editing while others \cite{nullinversion, edict} proposed improved inversion \cite{diffae} or train a new model \cite{instructpix2pix} specifically for text guided image editing.

Text guided image editing or image-to-image translation involves an input image, an edit text prompt and a trained model that edits the input image corresponding to the provided edit text/prompt while preserving structure and identity of the base image. Example edits from PRedItOR are shown in Fig.\ \ref{fig:intro}. A major constraint in most of the existing techniques \cite{prompt2prompt, imagic, unitune, plugnplay, edict} that are built on top of pretrained text-to-image DMs such as Stable Diffusion\footnote{https://stability.ai}, is the need for a base text prompt that corresponds to the input image. Such a base prompt is available for generated images but may not be available for real images. These techniques propose optimizations or fine-tuning of the model to arrive at a suitable base prompt or assume it is available to perform reasonable edits. Moreover, to the best of our knowledge, none of the diffusion model based editing techniques have looked at the DALLE-2 \cite{dalle2} architecture explicitly for text guided image editing, though it has been explored for video generation/stylization and image generation in \cite{dalle2, shifted_diffusion, runway, make_a_video}. 

In this paper, we propose a novel method to perform text guided image editing based on a pretrained text-to-image generation model that follows the DALLE-2 setup. Our proposed editing technique, \emph{PRedItOR} (EDiTing with PRIOR) does not require a base prompt \cite{prompt2prompt, imagic, unitune}, optimizations of embeddings \cite{imagic, unitune, edict, nullinversion}, fine-tuning model weights \cite{imagic}, additional guidance \cite{shape} or training new models and additional objectives \cite{instructpix2pix, plugnplay}. We observe that the Diffusion Prior model can be used to perform a text guided \emph{conceptual edit} of the base image's CLIP L/14 embedding. This is similar to performing a latent walk on the CLIP image embedding space by moving an embedding along a specific direction \cite{styleclip}. However, instead of manually discovering directions, the prior model automatically moves a base image embedding along a suitable direction determined by the text conditioning to generate an edit embedding. The generated CLIP embedding has context from both the edit text and the base image and a controllable parameter helps specify the trade-off between the contexts. The generated CLIP image embedding and the CLIP image embedding of the base image can then be used along with existing approaches like SDEdit \cite{sdedit} or Reverse DDIM \cite{ddim, diffae} to perform a \emph{structural edit} of the base image with the LDM Diffusion Decoder. 

To summarize, we propose the following key contributions in this paper:\\
-- Using the publicly available LAION prior model, we identify that the Diffusion Prior can be used to automatically make text guided conceptual edits in the CLIP image embedding space. \\
-- We train an LDM conditioned on normalized CLIP L/14 image embeddings to convert outputs from the LAION prior to images to support our experiments. \\
-- We utilize the Diffusion Prior's conceptual edit along with a structural edit on the Diffusion Decoder to perform controllable text guided image editing of an image using a pretrained HDM. Editing with PRedItOR is efficient and takes at most as much time as it takes for one text-to-image generation pipeline using the HDM. We perform comprehensive qualitative and quantitative analysis on PRedItOR and compare with existing baselines.\\
-- We highlight that PRedItOR provides an easy, intuitive, two-step controllable method to perform text guided image editing. It does not require expensive optimizations, finetuning, additional inputs, objectives or training. We show comparable or better performance than existing baselines and provide deeper analysis into the workings of the Diffusion Prior. 

\section{Related Works}
\label{sec:related}
\noindent\textbf{Diffusion Models:} DMs \cite{ddpm} are likelihood-based models that have gained popularity in image synthesis like Generative Adversarial Networks (GANs) \cite{diffusion_beat_gans, gan}. DMs are based on a Gaussian denoising process \cite{dickstein_2015} which assumes that the noises added to the original images are drawn from Gaussian distributions. The denoising process involves predicting the added noises using a Convolutional Neural Network (CNN) UNet \cite{unet}. Compared to GANs, DMs are easier to train and scale and have been shown to achieve state-of-the-art image quality \cite{diffusion_beat_gans, ddpm}. \\
\noindent\textbf{Conditional Diffusion Models:}
In conditional settings, DMs can be augmented with classifier guidance \cite{diffusion_beat_gans, glide}. The classifier guidance method allows DMs to condition on the predictions of a classifier. In classifier-free guidance \cite{classifier_free_guidance}, such classifiers are not needed. Instead, the guidance comes from the interpolation of the denoising network's outputs with and without conditioning. %

DMs can be conditioned on texts, images, or both \cite{ediff_i, muse, diffusion_beat_gans, classifier_free_guidance, improved_diffusion, glide, dalle2, ldm, imagen, parti}. DMs can also be applied to other computer vision tasks such as super-resolution \cite{glide, imagen} and in-painting \cite{repaint, sdedit, dalle2, palette}. Many techniques recently have improved DM's training and sampling speed. Instead of operating in the pixel space, Latent DMs \cite{ldm} are trained and sampled from a latent space \cite{vqgan} which is much smaller. Fast sampling methods \cite{k_diffusion, pndm, dpm_solver_pp, ddim, deis}, on the other hand, reduce the number of sampling steps. Some recent distillation techniques \cite{distillation_guided_diffusion, distillation_diffusion} have reduced the model complexity and sampling speed even further. We use LDM \cite{ldm} as the base for our decoder in contrast to DALLE-2 \cite{dalle2} for memory and compute efficiency. \\
\noindent\textbf{Diffusion Prior:} The Diffusion Prior was first introduced in OpenAI's DALLE-2 \cite{dalle2} which is a hierarchical DM. The diffusion prior uses a diffusion process much like the decoder and other diffusion models. However, it generates a CLIP image embedding from random noise, conditioned on CLIP text embedding and encodings using a Causal Transformer architecture \cite{dalle2}. Diffusion Decoder (unCLIP) then generates a synthetic image conditioned on the prior's output CLIP image embedding. In DALLE-2, the Diffusion Prior is shown to outperform an auto-regressive variant with respect to model size and training time. The Diffusion Prior is also used in \cite{make_a_video, runway} to train models using image only data but perform text to image/video during inference. Since the DALLE-2 model is not public and there are no variants of it available publicly, this hybrid architecture has not been explored as comprehensively as the direct text conditioned image generation architecture such as \cite{ldm, imagen}. 
Shifted Diffusion \cite{shifted_diffusion} recently proposed a novel Diffusion Prior architecture to improve relevance by sampling from a valid CLIP image embedding instead of random noise. However, to the best of our knowledge, there are no works that look at the characteristics and abilities of the Diffusion Prior, particularly for editing. \\
\noindent\textbf{Text guided Image Editing:} GANs have been well studied in the literature for image to image translation \cite{sae, munit, stargan, gan-translation-1}, latent manipulation and text guided editing \cite{stylegan, styleclip, stylespace, gan-edit-1, gan-edit-2, gan-edit-3, crowson2022vqgan, stylegannada, zheng2022bridging}. Many recent works have been proposed for text guided image editing on top of a base Stable Diffusion model. SDEdit \cite{sdedit} is an inference time technique for editing but its stochastic property makes it difficult to trade off between preserving structure of the base image and capturing the context of the edit prompt. Reverse DDIM \cite{ddim, diffae} helps with structure preservation but has limitations such as imperfect inversions and inaccurate predictions for classifier-free guidance (cfg) values greater than one \cite{nullinversion, edict}. It also assumes the base prompt associated with the original image is available. DiffusionCLIP \cite{diffusionclip} uses CLIP loss during inference to optimize the generation towards an edited image but involves an expensive optimization. 

\cite{diffusionclip, avrahami2022blended} are optimization based while Prompt-to-Prompt \cite{prompt2prompt} proposes a way to alter the cross-attention maps to replace objects or increase and decrease the effect of certain words and phrases in the generated image. Imagic \cite{imagic} and UniTune \cite{unitune} require expensive optimizations to obtain a reasonable base prompt corresponding to the base image. Imagic also requires fine-tuning model weights for improving inversions before using the optimized base prompt for editing. Null-inversion \cite{nullinversion} and EDICT \cite{edict} propose optimizations for improving inversions using Reverse DDIM for text conditioned image generation models like Stable Diffusion and show applications to editing. Plug-n-Play \cite{plugnplay} is a recent model that uses intermediate layers' attention maps to translate features from an image to another. All of the above techniques either require optimization to obtain a reasonable and accurate base prompt to ensure inversions are accurate for editing or assume the presence of such prompts. However, in reality only images generated by a text-to-image model have corresponding text that helps accurately invert the generated image. Inverting a real image on the other hand, requires conditional text and it is rather difficult to manually define appropriate text that would result in accurate inversions. This often results in inaccurate and subpar inversions requiring additional optimizations as described in \cite{nullinversion, edict}. 

PRedItOR overcomes the above limitations by using a Hybrid Diffusion Model where inversions are relatively more accurate (ref. Appendix Sec.\ref{appendix:inversion}) since the Diffusion Decoder is conditioned on CLIP image embeddings rather than text (ref. Sec. \ref{sec:structural}). The Diffusion Prior helps automatically edit the CLIP embedding to a reasonable edit embedding (ref. Sec. \ref{sec:conceptual}) that can be used for structure preserving edits in the decoder without requiring any base prompts or optimization. We believe this opens up interesting new questions and research directions with further applications of Diffusion Models for latent walks and multi-modal representation learning. 

\vspace{-0.1in}
\section{Proposed Method}
\label{sec:proposed}
The overall architecture of the diffusion model trained and used in this paper is similar to \cite{dalle2} with some changes for ease of implementation. Briefly, a Diffusion Prior model $\mathcal{P}_{\theta}$ generates a CLIP image embedding from random noise conditioned on a text prompt $y_e$. Since there are not publicly available weights or official code for the DALLE-2 model, we leverage the \emph{Diffusion Prior} model\footnote{https://github.com/LAION-AI/conditioned-prior} trained by LAION\footnote{https://laion.ai} that generates normalized CLIP L/14 embeddings conditioned on text for our experiments. The Diffusion Prior from LAION has been trained to generate a l2 normalized CLIP L/14 image embedding, given a text prompt.  The Diffusion Decoder $\mathcal{D}_{\phi}$ is a UNet based on LDM \cite{ldm} modified to take a single CLIP image embedding as conditioning to generate an image. Since we are not aware of publicly available DMs that generate images conditioned on normalized CLIP L/14 embeddings, we train a \emph{Diffusion Decoder} model based on \cite{ldm}. We use the pretrained VAEs from \cite{ldm} to convert VAE latents generated by the decoder to a pixel image. The entire model comprised of the Diffusion Prior and Diffusion Decoder is the Hybrid Diffusion Model (HDM) used for text-to-image generation. More information regarding the HDM is provided in the Appendix Sec.\ref{appendix:training}.

\begin{figure}
 \centering
 \includegraphics[width=\columnwidth]{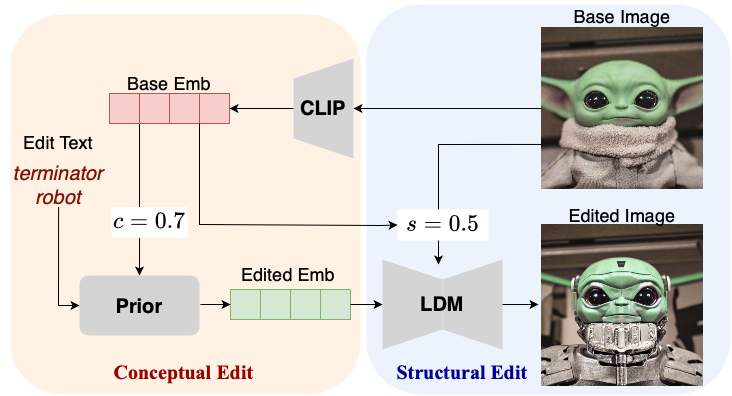}
 \captionof{figure}{The PRedItOR inference technique for text guided image editing comprising (a) conceptual edit of the CLIP image embedding of the base image using Diffusion Prior and (b) structural edit of the base image using CLIP embedding of the base image and the edited embedding from the Diffusion Prior. Both steps are controllable and can help generate a wide variety of possible edits as shown in Appendix Sec.\ \ref{appendix:examples}}
 \label{fig:preditor}
 \vspace{-0.2in}
\end{figure}
\subsection{PRedItOR}
\label{sec:preditor}
The proposed PRedItOR approach for text guided image editing is built on a pre-trained HDM model. It uses the Diffusion Prior to perform a \emph{conceptual edit} of the CLIP embedding of the base image $x_b$ followed by a \emph{structural edit} using the decoder. The overall process is depicted in Fig.\ \ref{fig:preditor}. PRedItOR is a inference time technique and the HDM along with VAE and CLIP encoders are frozen, as shown in Fig.\ \ref{fig:preditor}. The inputs are the base image $x_{b}$ and edit text/prompt $y_{e}$ and the output is the edited image $x_{e}$.
\vspace{-0.1in}
\subsubsection{Conceptual Editing}
\label{sec:conceptual}
In the \emph{Conceptual Edit} step, we use the Diffusion Prior model to modify the CLIP L/14 embedding $z_{b}$ of the input image $x_{b}$ to another embedding $z_{e}$ by conceptually imitating the process of moving the base embedding along the direction specified by the edit text $y_{e}$ in CLIP embedding space.

We discover that if $x_{b}$ is injected into the Diffusion Prior $\mathcal{P}_\theta$ at some intermediate timestep $t_c$ during sampling, and run the remaining sampling steps from $t_c$ to $0$ while conditioning on the CLIP text embedding $z_y$ (ref. Appendix Sec.\ref{appendix:proposed-prior} for further details) of the edit text prompt $y_{e}$, the resulting image embedding $z_{e}$ has context from both the base image and the edit prompt. This process is depicted in Eq.\ref{eq:conceptual_edit} where $p_\theta(x_{t-1} | x_t)$ depicts a single DDIM sampling step \cite{ddim, diffae, dalle2} starting from timestep $t_c$ instead of T and $z_{t_{c}}$ = $z_b$. 

\begin{equation}
p_{\theta}(z_{e} | z_b, z_y) = \prod_{t=1}^{t_c} p_\theta(z_{t-1} | z_{t}, z_y)
\label{eq:conceptual_edit}
\end{equation}

The higher the value of $t_c$, the more number of steps the prior gets to modify the injected embedding according to edit text and the closer the generated embedding will be to the edit text. We provide an intuitive illustration of the proposed \emph{conceptual edit} in Sec.\ref{appendix:shifted}. We also show the generated image embedding's CLIP similarity to the input base image and edit prompt as a function of the injection timestep $t_{s}$ in Appendix Sec.\ \ref{appendix:prior_visual}. 
We control the injection timestep $t_c$ using a \emph{conceptual edit} strength parameter $c \in [0, 1]$ and $t_c$ = T $\times$ $c$. The higher the value of $c$, the more the base embedding $z_b$ will be modified. More analyses on the prior's intermediate outputs and intuition are provided in Sec.\ \ref{appendix:prior_visual}.
\vspace{-0.1in}
\subsubsection{Structural Editing }
\label{sec:structural}
The above conceptual edit step gives us a CLIP image embedding that has context from both the base image $x_b$ and edit text $y_e$. To generate the final edited image $x_e$, we pass the generated edit embedding $z_e$ to the Diffusion Decoder. However, just passing the embedding through the decoder will generate an image that may not preserve the structure of the base image due to the lack of spatial information in the generated CLIP embedding \cite{clip, dalle2}. Since the noise map $\mathbf{z_T}$ sampled from standard normal is leveraged to fix structure within the first few steps of the reverse diffusion process, we can perform structure preserving edits by starting from noise that deterministically generates the base image and modify the conditioning at later timesteps. We use the Reverse DDIM \cite{ddim, diffae} technique to get the noise by deterministically running the reverse of the reverse diffusion process on $\mathbf{z_0}$ conditioned on the base image's CLIP embedding $z_b$. If $\mathbf{z_t}$ represents the noised VAE latent at some timestep $t$, $\epsilon_{\theta}(\mathbf{z_t}, t, z_b)$ is the noise prediction decoder UNet, and $\mathbf{f}_{\theta}(\mathbf{z_t}, t, z_b)$ is parameterized by the noise prediction network as $\frac{1}{\sqrt{\alpha_t}}(\mathbf{z_t} - \sqrt{1 - \alpha_{t}}\epsilon_{\theta}(\mathbf{z_t}, t, z_b))$, then a single step of the reverse DDIM process with our Diffusion Decoder can be depicted as  in Eq.\ref{eq:structural_edit} following \cite{ddim, diffae}.

\begin{equation}
\mathbf{z_{t+1}} = \sqrt{\alpha_t + 1} \cdot \mathbf{f}_{\theta}(\mathbf{z_t}, t, z_b) + \sqrt{1 - \alpha_{t+1}} \cdot \epsilon_{\theta}(\mathbf{z_t}, t, z_b)
\label{eq:structural_edit}
\end{equation}

The reverse DDIM process starts from the base image's VAE latent $\mathbf{z_0}$=VAE$_{enc}(x_b)$. We use a \emph{structural edit} strength parameter $s \in [0, 1]$ and $t_s$ = T $\times$ $s$ that controls the strength of structure preservation by specifying the timestep $t_s$ until which the reverse DDIM process is run. The larger the strength parameter $s$, the larger the timestep $t_s$ and noisier the base image becomes. Hence, increasing $s$ corresponds to reducing the preserved structural information. %
Once we get the deterministic noised latent $\mathbf{z_{t_{s}}}$, we run the regular sampling process of the diffusion decoder starting from $t_s$ until $0$, conditioned on the \emph{conceptually edited} embedding  $z_e$ to get the final edited latent $\mathbf{\hat{z}_e}$. The generated VAE latent can then be passed through the pre-trained and fixed VAE decoder to get the final edited image $x_e$ as $x_e$=VAE$_{dec}(\mathbf{\hat{z}_e})$. It is also possible to apply this \emph{structural edit} step using SDEdit \cite{sdedit}, however, SDEdit is stochastic and has been observed \cite{nullinversion, edict} to not preserve structure as effectively for larger strengths. We show similar results in Sec.\ \ref{sec:ablation}. 

\begin{figure*}[t]
 \begin{tabular}{c@{\hspace{0.5cm}}c@{\hspace{0.1cm}}c@{\hspace{0.1cm}}c@{\hspace{0.1cm}}c@{\hspace{0.1cm}}c@{\hspace{0.1cm}}c@{}}
 Input & Prior(rDDIM) & Prior(SDEdit) & SD(0.3)& SD(0.6)& Ours(SDEdit)& Ours(rDDIM) \\ 
 \midrule
 \rotatebox[x=1mm,y=6mm]{90}{\emph{fluffy}}\hspace{0.04in}
 \includegraphics[width=.13\linewidth]{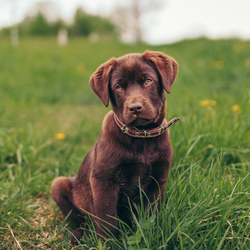}&
 \includegraphics[width=.13\linewidth]{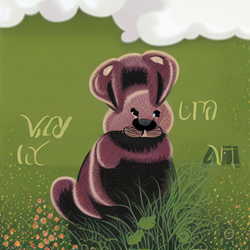} &%
 \includegraphics[width=.13\linewidth]{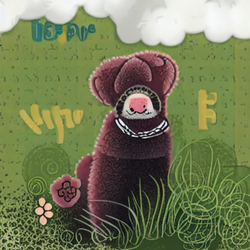} &%
 \includegraphics[width=.13\linewidth]{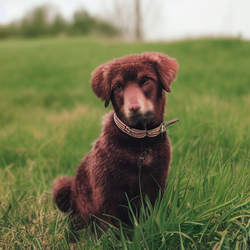} &
 \includegraphics[width=.13\linewidth]{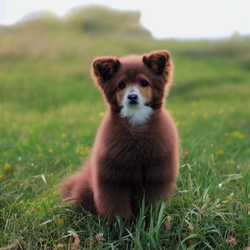} &
 \includegraphics[width=.13\linewidth]{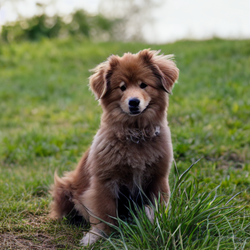} &
 \includegraphics[width=.13\linewidth]{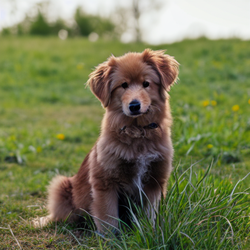} \\
 \rotatebox[x=1mm,y=5mm]{90}{\emph{tomatoes}}\hspace{0.05in}
 \includegraphics[width=.13\linewidth]{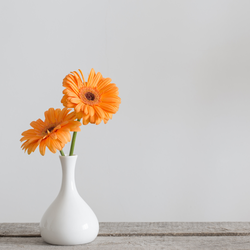}&
 \includegraphics[width=.13\linewidth]{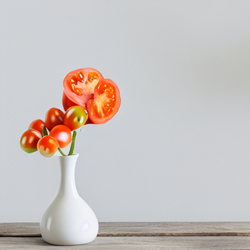} &%
 \includegraphics[width=.13\linewidth]{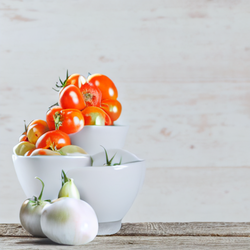} &%
 \includegraphics[width=.13\linewidth]{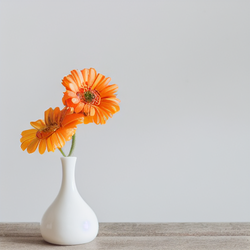} &
 \includegraphics[width=.13\linewidth]{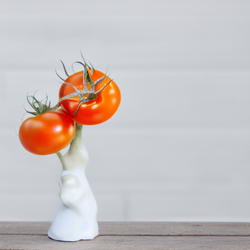}&
 \includegraphics[width=.13\linewidth]{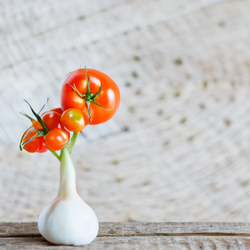} &
 \includegraphics[width=.13\linewidth]{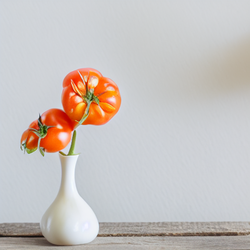} \\
 \rotatebox[x=1mm,y=1mm]{90}{\emph{basket of kittens}}\hspace{0.05in}
 \includegraphics[width=.13\linewidth]{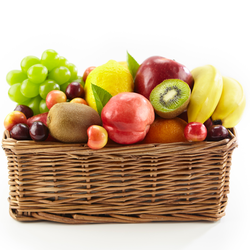}&
 \includegraphics[width=.13\linewidth]{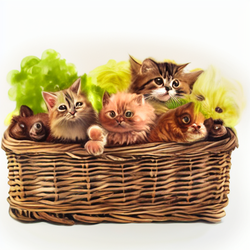}&%
 \includegraphics[width=.13\linewidth]{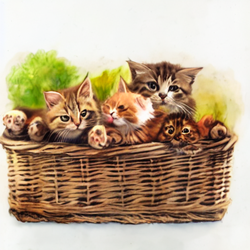}&%
 \includegraphics[width=.13\linewidth]{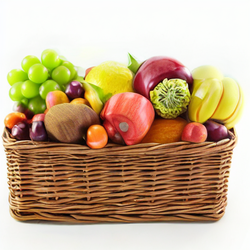}&
 \includegraphics[width=.13\linewidth]{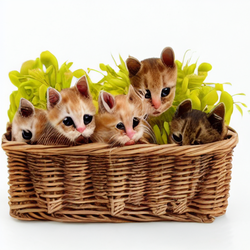}&
 \includegraphics[width=.13\linewidth]{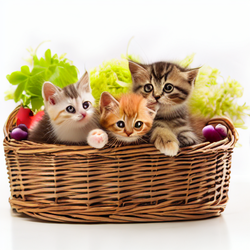}&
 \includegraphics[width=.13\linewidth]{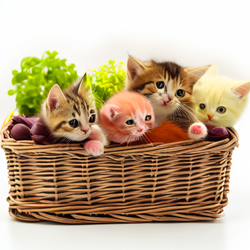} \\
 \rotatebox[x=1mm,y=8mm]{90}{\emph{robot}}\hspace{0.05in}
 \includegraphics[width=.13\linewidth]{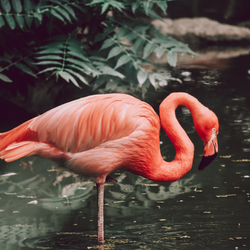}&
 \includegraphics[width=.13\linewidth]{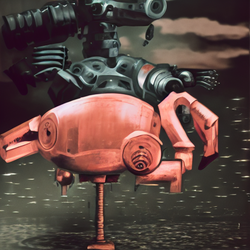}&
 \includegraphics[width=.13\linewidth]{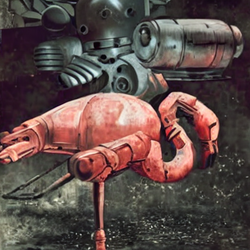}&
 \includegraphics[width=.13\linewidth]{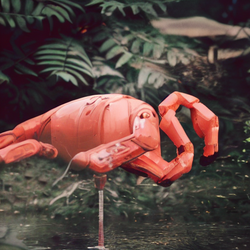}&
 \includegraphics[width=.13\linewidth]{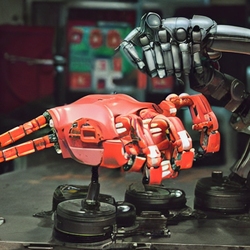}&
 \includegraphics[width=.13\linewidth]{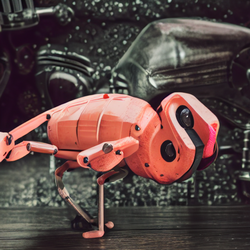} &
 \includegraphics[width=.13\linewidth]{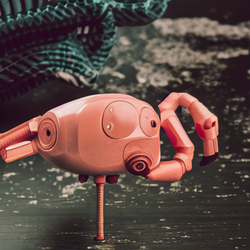} 
 
\end{tabular}
 \caption{Qualitative examples showing comparison with ablated/alternate versions and SDEdit on Stable Diffusion (SD). SD($s$) refers to applying SDEdit on Stable Diffusion with strength $s$. We can clearly see that the proposed PRedItOR performs the most plausible edits while preserving the background and structure of the base image.}
 \label{fig:ablations}
 \vspace{-0.1in}
\end{figure*} 
\section{Experiments and Results}
\label{sec:experiments}
For text guided image editing, we show results from the proposed PRedItOR setup in Fig.\ \ref{fig:intro}.\footnote{We show general text-to-image generation results using our custom LDM decoder and LAION prior in the Appendix Sec.\ref{appendix:hdm} since that is not the focus of this paper. }\ We perform a comprehensive qualitative comparison with existing baselines as well as ablations and alternatives for the proposed setup. We also perform quantitative comparison using CLIP score for relevance and LPIPS score for structural similarity and provide results comparing with baselines in Appendix Sec.\ref{appendix:quant}. %

Our full proposed setup of \emph{conceptual edit} followed by \emph{structural edit} is referred to  as PRedItOR. The \emph{structural edit} step can be performed by either applying the reverse DDIM process described in Sec.\ref{sec:structural} or SDEdit \cite{sdedit}. These variants are identified as Ours(rDDIM) and Ours(SDEdit) respectively. 

\subsection{Ablation Study}
\label{sec:ablation}
We start by ablating the primary contribution of this paper, the \emph{conceptual edit} step using Diffusion Prior. \\
\noindent\textbf{Prior(rDDIM):} To show the importance and effectiveness of the proposed \emph{conceptual edit} step, we show results without this step. We pass the edit text to the Diffusion Prior to generate an image embedding and then apply the \emph{structural edit} step normally. \\
\noindent\textbf{Prior(SDEdit):} This is the same as Prior(rDDIM) but applies SDEdit during the \emph{structural edit} step.\\
\noindent\textbf{SD \cite{ldm}(SDEdit):} This is Stable Diffusion with SDEdit using the edit text.\\
\noindent\textbf{PRedItOR(SDEdit):} This is the proposed two step PRedItOR process with SDEdit for \emph{structural edit}. This could also be referred to as Ours(SDEdit).\\
\noindent\textbf{PRedItOR:} This is the proposed two step PRedItOR process with reverse DDIM for \emph{structural edit}. Note that PRedItOR, PRedItOR(rDDIM), Ours and Ours(rDDIM) are all equivalent unless otherwise mentioned.

\subsection{Existing Baselines}
\label{sec:baselines}
We compare with the existing text guided image editing baselines shown in Table.\ref{tab:baselines}. We emphasize the characteristics that differentiate the proposed technique from the baselines in the last two columns. Almost all baselines either require a carefully designed base prompt, optimization of embedding, finetuning of existing model weights or training new models like \cite{instructpix2pix}. Our proposed method, assuming a pre-trained DALLE-2 like model exists, can be applied without any of the additional overhead. 
\begin{table}[!h]
 \centering
 \begin{tabular}{p{0.25\linewidth}p{0.2\linewidth}p{0.15\linewidth}p{0.2\linewidth}}
 \toprule
 Baseline & Base model & Requires base prompt & Requires optimization or finetuning\\
 \midrule 
 P2P \cite{prompt2prompt}&SD&\checkmark&$\times$ \\
 Imagic \cite{imagic}&SD&\checkmark&\checkmark \\
 UniTune \cite{unitune}&Imagen&\checkmark&\checkmark \\
 EDICT \cite{edict}&SD&\checkmark&\checkmark \\
 P\&P \cite{plugnplay}&SD&\checkmark&$\times$ \\
 SGD \cite{shape}&SD&\checkmark&$\times$ \\
 IP2P \cite{instructpix2pix}&--&$\times$&\checkmark \\
 \rowcolor[gray]{0.9}
 PRedItOR&DALLE-2&$\times$&$\times$ \\
 \bottomrule
 \end{tabular}
 \caption{Baseline methods used for comparison.}
 \label{tab:baselines}
 \vspace{-0.1in}
\end{table}
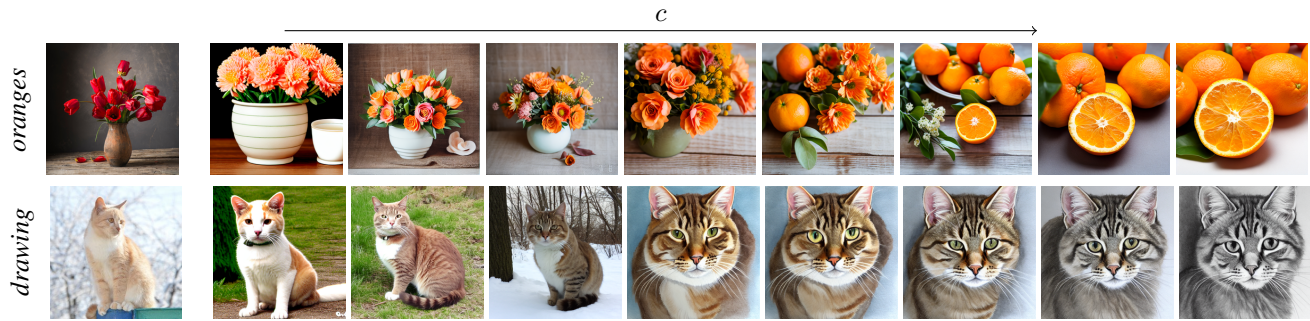
\begin{figure*}
 \centering
 \begin{tikzpicture}
 \draw [->] (0,0) -- (10,0) node [midway, above] {$c$};
\end{tikzpicture} \\ \vspace{0.05in} 
 \rotatebox[x=1mm,y=4mm]{90}{\emph{oranges}}\hspace{0.05in}
 \includegraphics[width=.1\linewidth]{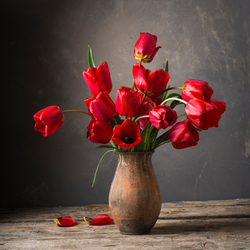} \hspace{0.1in}
 \includegraphics[width=.1\linewidth]{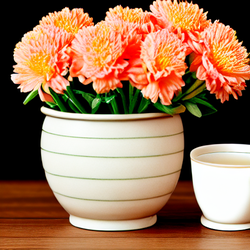}
 \includegraphics[width=.1\linewidth]{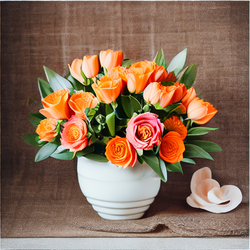}
 \includegraphics[width=.1\linewidth]{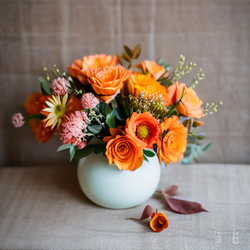}
 \includegraphics[width=.1\linewidth]{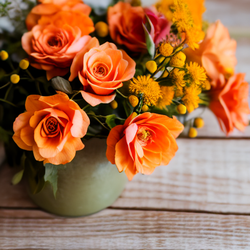}
 \includegraphics[width=.1\linewidth]{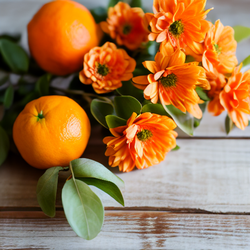}
 \includegraphics[width=.1\linewidth]{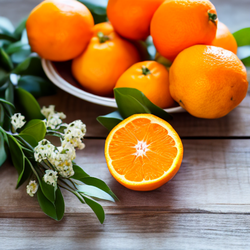}
 \includegraphics[width=.1\linewidth]{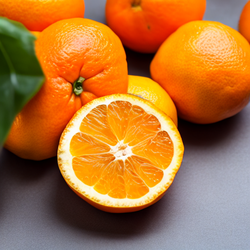}
 \includegraphics[width=.1\linewidth]{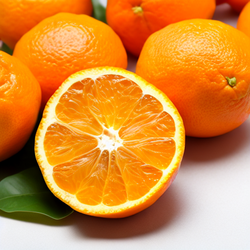}
 \\ \vspace{0.05in} 
 \rotatebox[x=1mm,y=4mm]{90}{\emph{drawing}}\hspace{0.05in}
 \includegraphics[width=.1\linewidth]{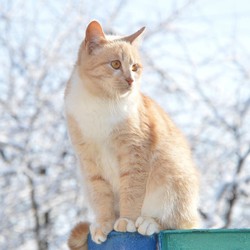} \hspace{0.1in}
 \includegraphics[width=.1\linewidth]{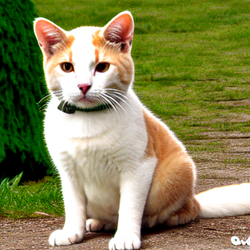}
 \includegraphics[width=.1\linewidth]{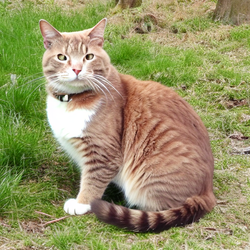}
 \includegraphics[width=.1\linewidth]{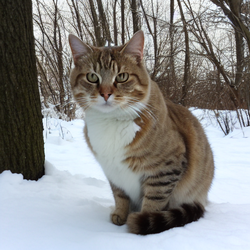}
 \includegraphics[width=.1\linewidth]{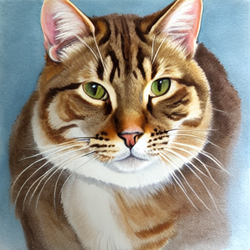}
 \includegraphics[width=.1\linewidth]{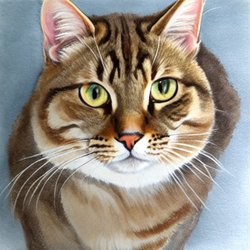}
 \includegraphics[width=.1\linewidth]{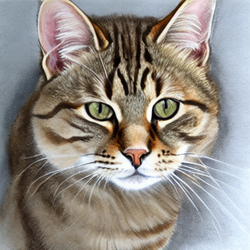}
 \includegraphics[width=.1\linewidth]{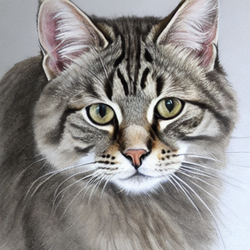}
 \includegraphics[width=.1\linewidth]{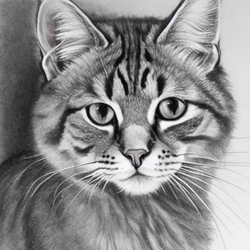}
 \caption{Visualizing the embedding generated by the \emph{conceptual edit} step for two samples. We can see that without the edit text mentioning the cat, the sketch obtained was that of a cat as a result of initializing the refinement with the base cat image. Similarly, the concept of `oranges' is gradually encoded into the embedding as $c$ increases.}
 \vspace{-0.1in}
 \label{fig:conceptual}
\end{figure*}

For fair comparison, instead of running their codes (some of which are not available publicly) on new images with edit text, we apply our method on the examples from the respective papers. We get the base images from the papers or project websites. This way, our results on their methods do not depend on the hyperparameters that we choose. We show a few examples in the main paper and include more examples along with quantitative evaluation in the Appendix Sec.\ref{appendix:quant}. 
\begin{figure*}[ht]
 \begin{tabular}{p{2mm}p{3mm}c@{\hskip 0.4mm}c@{\hskip 0.4mm}c@{\hskip 3mm}p{3mm}c@{\hskip 0.4mm}c@{\hskip 0.4mm}c}
 \arrayrulecolor{gray}
 & $y_e$& $x_b$& Baseline & PRedItOR & $y_e$ & $x_b$& Baseline & PRedItOR\\ \midrule
 
 \rotatebox[x=1mm,y=-7mm]{90}{P2P \cite{prompt2prompt}}&
 \rotatebox[x=1mm,y=-7mm]{90}{\emph{flooded street}}&
 \raisebox{-.5\height}{\includegraphics[width=.15\linewidth]{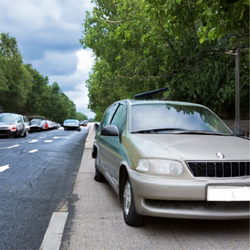}} &%
 \raisebox{-.5\height}{\includegraphics[width=.15\linewidth]{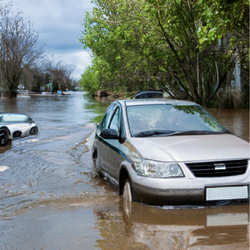}} &%
 \raisebox{-.5\height}{\includegraphics[width=.15\linewidth]{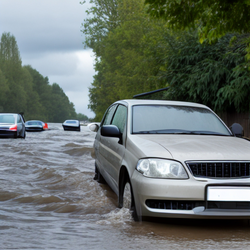}} &

 \rotatebox[x=1mm,y=-11mm]{90}{\emph{basket of oranges}}&
 \raisebox{-.5\height}{\includegraphics[width=.15\linewidth]{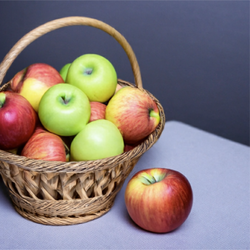}} &%
 \raisebox{-.5\height}{\includegraphics[width=.15\linewidth]{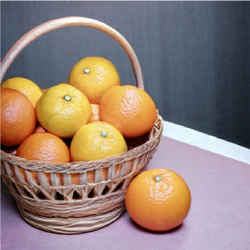}} &%
 \raisebox{-.5\height}{\includegraphics[width=.15\linewidth]{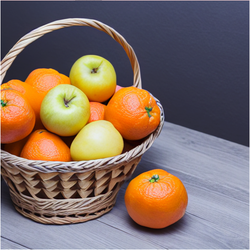}} \\ \midrule 

 \rotatebox[x=1mm,y=-7mm]{90}{Imagic \cite{imagic}}&
 \rotatebox[x=1mm,y=-8mm]{90}{\emph{pistachio cake}}&
 \raisebox{-.5\height}{\includegraphics[width=.15\linewidth]{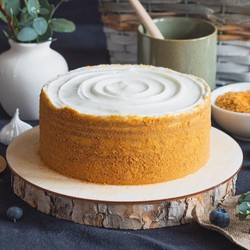}} &%
 \raisebox{-.5\height}{\includegraphics[width=.15\linewidth]{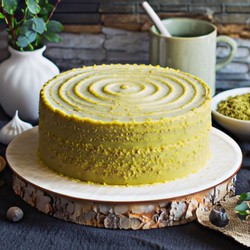}} &%
 \raisebox{-.5\height}{\includegraphics[width=.15\linewidth]{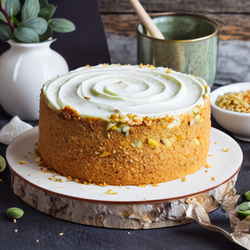}} &
 \rotatebox[x=1mm,y=-4mm]{90}{\emph{zebra}}&
 \raisebox{-.5\height}{\includegraphics[width=.15\linewidth]{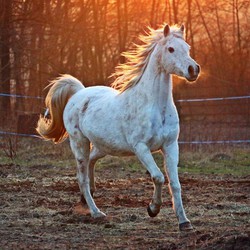}} &%
 \raisebox{-.5\height}{\includegraphics[width=.15\linewidth]{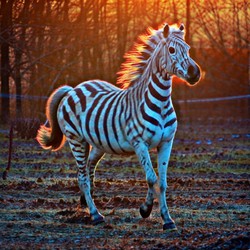}} &%
 \raisebox{-.5\height}{\includegraphics[width=.15\linewidth]{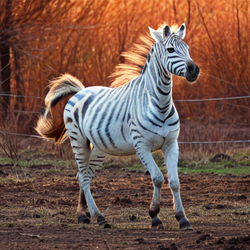}} \\ \midrule 

 \rotatebox[x=1mm,y=-7mm]{90}{UniTune \cite{unitune}}&
 \rotatebox[x=1mm,y=-11mm]{90}{\emph{watercolor painting}}&
 \raisebox{-.5\height}{\includegraphics[width=.15\linewidth]{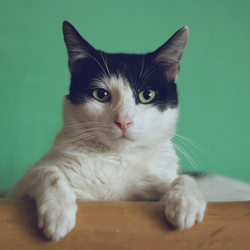}} &%
 \raisebox{-.5\height}{\includegraphics[width=.15\linewidth]{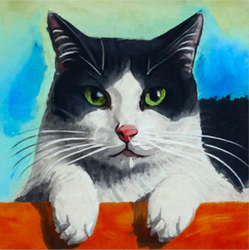}} &%
 \raisebox{-.5\height}{\includegraphics[width=.15\linewidth]{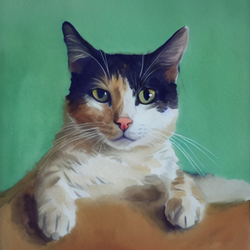}} &
 \rotatebox[x=1mm,y=-10mm]{90}{\parbox{2cm}{\emph{ladybug on a flower}}}&
 \raisebox{-.5\height}{\includegraphics[width=.15\linewidth]{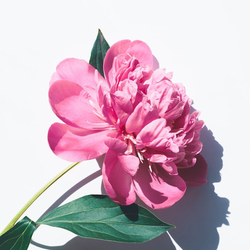}} &%
 \raisebox{-.5\height}{\includegraphics[width=.15\linewidth]{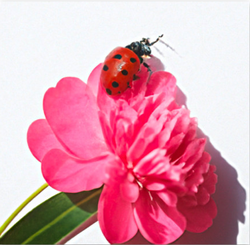}} &%
 \raisebox{-.5\height}{\includegraphics[width=.15\linewidth]{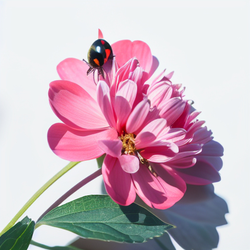}} \\ \midrule 

 \rotatebox[x=1mm,y=-7mm]{90}{EDICT \cite{edict}}&
 \rotatebox[x=1mm,y=-10mm]{90}{\parbox{2cm}{\emph{red chair at grand canyon}}}&
 \raisebox{-.5\height}{\includegraphics[width=.15\linewidth]{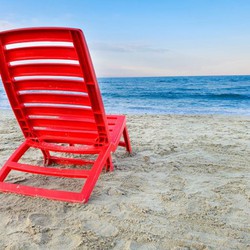}} &%
 \raisebox{-.5\height}{\includegraphics[width=.15\linewidth]{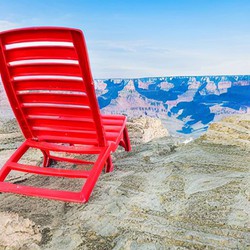}} &%
 \raisebox{-.5\height}{\includegraphics[width=.15\linewidth]{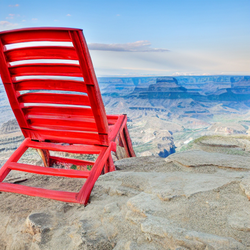}} &
 \rotatebox[x=1mm,y=-10mm]{90}{\emph{golden retriever}}&
 \raisebox{-.5\height}{\includegraphics[width=.15\linewidth]{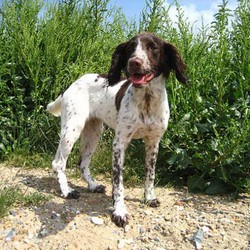}} &%
 \raisebox{-.5\height}{\includegraphics[width=.15\linewidth]{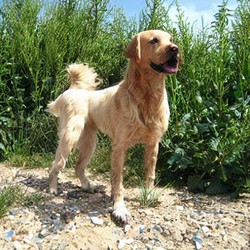}} &%
 \raisebox{-.5\height}{\includegraphics[width=.15\linewidth]{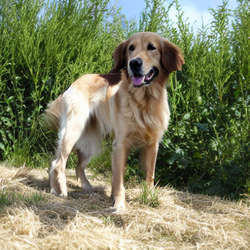}} \\ \midrule 

 \rotatebox[x=1mm,y=-7mm]{90}{PNP \cite{plugnplay}}&
 \rotatebox[x=1mm,y=-10mm]{90}{\parbox{2.5cm}{\emph{a photorealistic cat and bunny}}}&
 \raisebox{-.5\height}{\includegraphics[width=.15\linewidth]{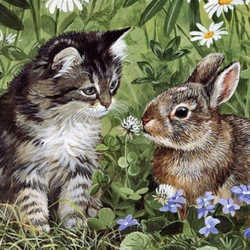}} &%
 \raisebox{-.5\height}{\includegraphics[width=.15\linewidth]{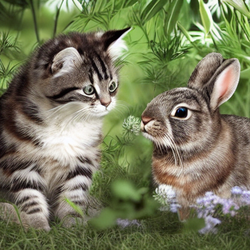}} &%
 \raisebox{-.5\height}{\includegraphics[width=.15\linewidth]{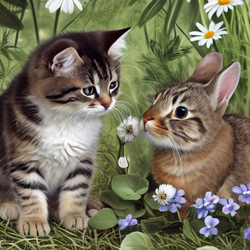}} &
 \rotatebox[x=1mm,y=-10mm]{90}{\parbox{2cm}{\emph{a cartoon of a jeep}}}&
 \raisebox{-.5\height}{\includegraphics[width=.15\linewidth]{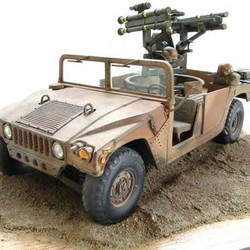}} &%
 \raisebox{-.5\height}{\includegraphics[width=.15\linewidth]{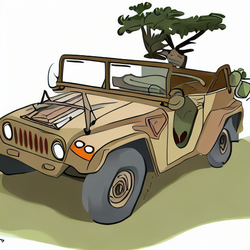}} &%
 \raisebox{-.5\height}{\includegraphics[width=.15\linewidth]{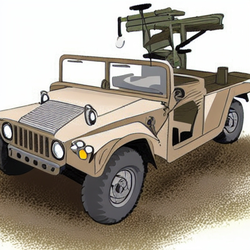}} \\ \midrule 
 
 \rotatebox[x=1mm,y=-7mm]{90}{SGD \cite{shape}}&
 \rotatebox[x=1mm,y=-10mm]{90}{\emph{stuffed bear}}&
 \raisebox{-.5\height}{\includegraphics[width=.15\linewidth]{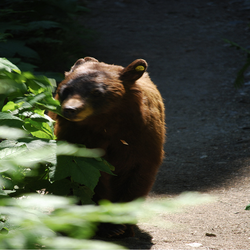}} &%
 \raisebox{-.5\height}{\includegraphics[width=.15\linewidth]{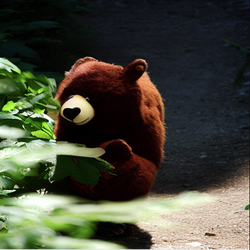}} &%
 \raisebox{-.5\height}{\includegraphics[width=.15\linewidth]{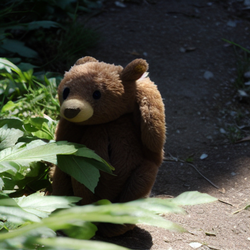}} &
 \rotatebox[x=1mm,y=-10mm]{90}{\emph{lego truck}}&
 \raisebox{-.5\height}{\includegraphics[width=.15\linewidth]{images/baselines/sgd/truck.png}} &%
 \raisebox{-.5\height}{\includegraphics[width=.15\linewidth]{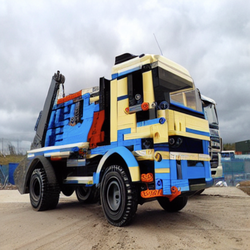}} &%
 \raisebox{-.5\height}{\includegraphics[width=.15\linewidth]{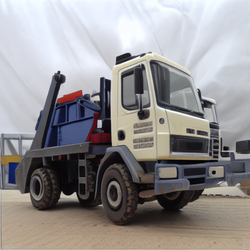}} \\ \midrule 

 \rotatebox[x=1mm,y=-7mm]{90}{IP2P \cite{instructpix2pix}}&
 \rotatebox[x=1mm,y=-5mm]{90}{\emph{roses}}&
 \raisebox{-.5\height}{\includegraphics[width=.15\linewidth]{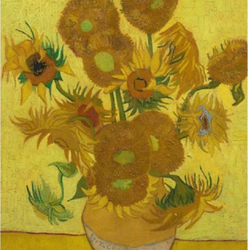}} &%
 \raisebox{-.5\height}{\includegraphics[width=.15\linewidth]{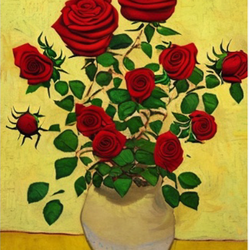}} &%
 \raisebox{-.5\height}{\includegraphics[width=.15\linewidth]{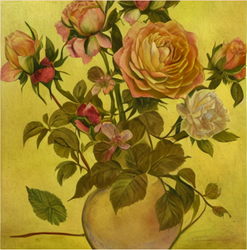}} &
 \rotatebox[x=1mm,y=-10mm]{90}{\emph{cherry blossoms}}&
 \raisebox{-.5\height}{\includegraphics[width=.15\linewidth]{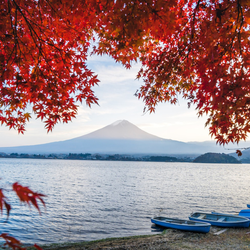}} &%
 \raisebox{-.5\height}{\includegraphics[width=.15\linewidth]{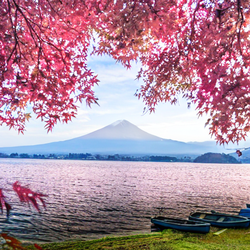}} &%
 \raisebox{-.5\height}{\includegraphics[width=.15\linewidth]{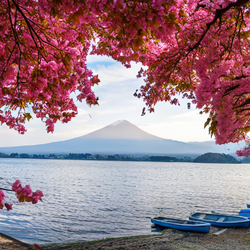}} \\ \midrule 
 \end{tabular}
 \caption{Qualitative examples comparing the proposed PRedItOR approach with existing baselines for text guided image editing. We can observe that the proposed setup achieves as good as or better than baselines in most cases. More importantly, PRedItOR does not require manually curated base prompt, edit prompt, optimization, finetuning or model training and performs structure preserving edits.}
 \label{fig:baselines}
 \vspace{-1.5mm}
\end{figure*}
\section{Results and Discussion}
\label{sec:results}
\noindent\textbf{The Effect of Conceptual Refinement:} From Fig.\ \ref{fig:ablations}, we can see that SDEdit's stochastic noise addition on Stable Diffusion, changes most of the background and structure for all examples. Moreover, without using the proposed \emph{conceptual edit} step (ref. Prior(SDEdit) and Prior(rDDIM)), the edit text `Rottweiler' does not have information that the required embedding needs to be of a realistic photograph. The prior's bias towards oil paintings gets reflected in the generated embedding. However, when the \emph{conceptual edit} is used  (ref.\ Ours(SDEdit) and Ours(rDDIM)), the edited embedding obtained from the base image, captures the concept that the base image is realistic. As a result, without carefully designed edit prompt, we get the necessary edit while preserving structure and background information from the original image. This shows the importance of the \emph{conceptual edit} step from the prior. \\
\noindent\textbf{Visualizing the Conceptual Edits:} We believe that the \emph{conceptual edit} step in the prior imitates a text guided latent walk. The prior is a regular diffusion model with a denoising network. But since it works in the CLIP embedding space, one could imagine that `noise' is `conceptual' in nature rather than visual. Hence, when we inject the CLIP embedding of a base image at some intermediate step $t$, it is `noisy' with respect to the edit text given as conditioning. During the sampling process, the prior refines the injected embedding by denoising it to associate with the provided edit text. We visualize the output of \emph{conceptual edit} step and the effect of the strength parameter $c$, introduced in Sec.\ \ref{sec:conceptual} in Fig.\ \ref{fig:conceptual}. We can see that as $c$ increases, the number of denoising steps increases and the generated embedding is dominated by the edit text. To generate these images, we pass the refined embedding $z_{e}$ to the LDM decoder for regular sampling without the structural edit step, effectively visualizing the information in the generated embedding. \\
\noindent\textbf{Baseline Comparisons:} We apply our proposed PRedItOR setup using the pre-trained HDM on randomly chosen examples from all baselines provided in Tab.\ \ref{tab:baselines}. Results are shown in Fig.\ \ref{fig:baselines}. We can see that the proposed method performs as well as or better than baselines in most cases. We believe that our setup benefits from the controllable \emph{conceptual edit} setup using the Diffusion Prior and more accurate inversions in the \emph{structural edit} setup as a result of the Diffusion Decoder being conditioned only on CLIP image embedding. For example, in the first example of the flooded street, the proposed method does not change the design of the car as much as the P2P \cite{prompt2prompt} baseline. Similarly in the \emph{watercolor painting} example comparing UniTune \cite{unitune}, the pose, size and appearance of the cat remains unchanged in our example while the style changes according to the edit text. Similar observations can be made from other examples as well.
\begin{figure}[t]
 \centering
 \hspace{.15\linewidth} Without mask \hspace{0.05in} With mask\\ \vspace{0.1in}
 \rotatebox[x=1mm,y=4mm]{90}{\emph{a husky}}\hspace{0.05in}
 \includegraphics[width=.25\linewidth]{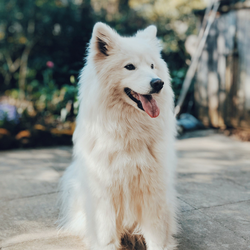} \hspace{0.05in}
 \includegraphics[width=.25\linewidth]{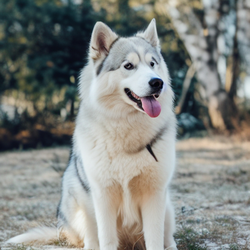}
 \includegraphics[width=.25\linewidth]{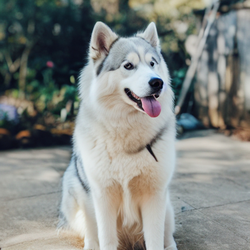}
 \fbox{\includegraphics[width=.1\linewidth]{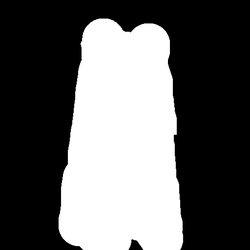}}\\ 
 \vspace{0.05in} 
 \rotatebox[x=1mm,y=3mm]{90}{\emph{in snow}}\hspace{0.05in}
 \includegraphics[width=.25\linewidth]{images/mask/masked_a_base.png} \hspace{0.05in}
 \includegraphics[width=.25\linewidth]{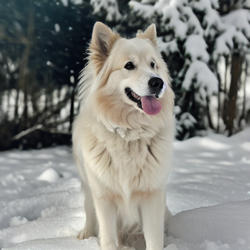}
 \includegraphics[width=.25\linewidth]{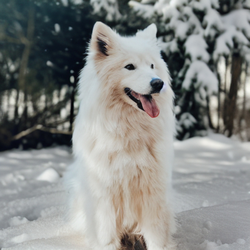}
 \fbox{\includegraphics[width=.1\linewidth]{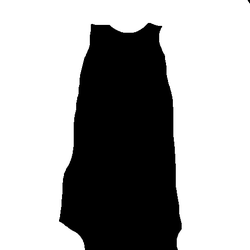}}
 \caption{Leveraging a mask as described in \cite{diffedit} during the \emph{structural edit} step can further enhance the results. The last column shows edits with the mask shown next to them. These edits ensure the unmasked regions are retained as they are in the base image. Base image obtained from Unsplash.}
 \vspace{-0.2in}
 \label{fig:masking}
\end{figure} \\
\noindent\textbf{Optional Mask Guidance:} For effective editing of a localized region of a base image without modifying other regions in the image, we could also use the technique proposed in DiffEdit \cite{diffedit} for better structure and background preservation. We show some examples in Fig.\ \ref{fig:masking} where adding an optional mask during \emph{structural edit} step enhances the results by leaving the unmasked (black) regions untouched. In the \emph{husky} example, changing the dog to a husky inadvertently adds snow to the background. Though these results are still better than existing baselines (ref.\ Sec.\ \ref{sec:baselines}) in terms of structure preservation, we find that performing the \emph{structural edit} only on the masked (white) region as described in \cite{diffedit} ensures that even the slightest changes to the unmasked regions are avoided. More examples and analysis are provided in Appendix Sec.\ \ref{appendix:examples}. \\ %
\noindent\textbf{Effect of Hyperparameters:} In Fig.\ \ref{fig:hyperparameters} we show the effect of the \emph{conceptual edit} strength $c$ and the \emph{structural edit} strength $s$ on the final edits. Increasing $c$ for the same $s$ make the edits more pronounced while increasing $s$ changes the overall structure of the base image. We find that it is intuitive to arrive at optimal values of $c$ and $s$ for an edit. One can also visualize the edited embedding after the \emph{conceptual edit} step as shown in Fig.\ \ref{fig:conceptual} without \emph{structural edit} to make an informed decision. In our experience, $c$ between 0.55 and 0.8 and $s$ between 0.3 and 0.6 give optimal results for most examples. If a mask is used, then $s$ can be increased further, since changes to unmasked regions are protected while using the DiffEdit\cite{diffedit} approach for masked editing.
\begin{figure}[t]
 \centering
 \begin{tabular}{c@{\hspace{0.1cm}}c@{\hspace{0.1cm}}c@{\hspace{0.1cm}}}
 \centering
 $s$=0.40&$s$=0.55&$s$=0.70\\
 \rotatebox[x=1mm,y=8mm]{90}{$c$=0.50}
 \includegraphics[width=.3\linewidth]{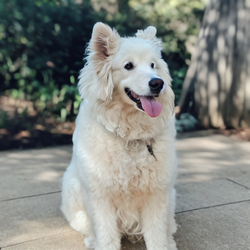}&
 \includegraphics[width=.3\linewidth]{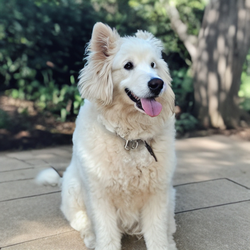}&
 \includegraphics[width=.3\linewidth]{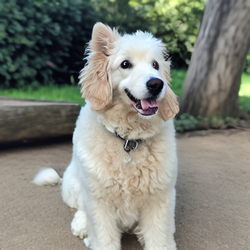} \\
 \rotatebox[x=1mm,y=8mm]{90}{$c$=0.65}
 \includegraphics[width=.3\linewidth]{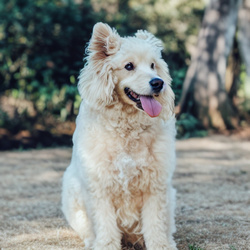}&
 \includegraphics[width=.3\linewidth]{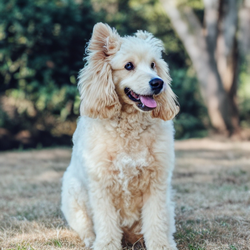}&
 \includegraphics[width=.3\linewidth]{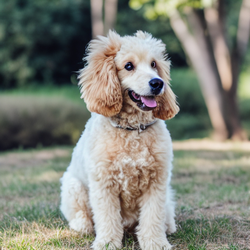} \\
 \rotatebox[x=1mm,y=8mm]{90}{$c$=0.80}
 \includegraphics[width=.3\linewidth]{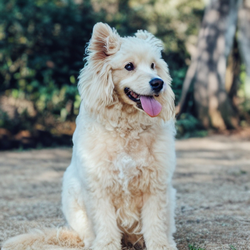}&
 \includegraphics[width=.3\linewidth]{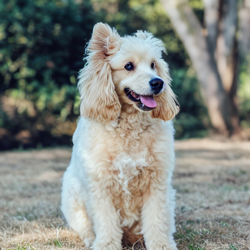}&
 \includegraphics[width=.3\linewidth]{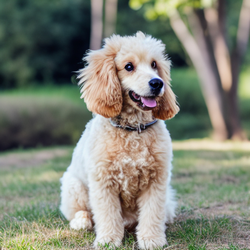}\\
 \end{tabular}
 \caption{The effect of controllable \emph{conceptual edit} and \emph{structural edit} strength parameters $c$ and $s$ respectively are shown. As $c$ increases, we see features of a `poodle' are better reflected conceptually in the edited image whereas as $s$ increases, more of the image is modified including the pose of the dog and the background. The base image for this example is from Fig.\ \ref{fig:masking}}
 \vspace{-0.2in}
 \label{fig:hyperparameters}
\end{figure}
\section{Conclusion}
\label{sec:discussion}
Through this work, we show that Diffusion Prior can be used to perform text guided conceptual editing (imitating a latent walk) in the CLIP embedding space, intuitively similar to the explicitly formulated shifted diffusion \cite{shifted_diffusion}. %
We identify that the \emph{Diffusion Prior} model can be used for text guided \emph{conceptual editing} in the CLIP embedding space. This is more intuitive and natural compared to approaches like StyleCLIP \cite{styleclip} that involves manual curation of positive and negative text to get the mean difference as a direction to guide a CLIP embedding. We then perform \emph{structural edit} using a CLIP image embedding conditioned decoder that natively performs better inversions without the need for optimizations such as Null inversion \cite{nullinversion} or EDICT \cite{edict}. We show controllable, structure preserving edits on various images and text prompts and compare with baselines and ablations to show the efficacy of the proposed technique. Our approach does not rely on manually defined base prompts, finetuning of embeddings, optimization of model weights or additional objectives, or additional inputs. We believe PRedItOR opens up interesting new research directions for Diffusion Prior as a learned automated way for conceptual manipulation. 

{\small
\bibliographystyle{ieee_fullname}
\bibliography{egbib}
}
\clearpage
\appendix

\begin{center}
\Huge{Appendix}    
\end{center}
\vspace{0.2in}
\normalsize
\section{Introduction}
\label{appendix:intro}
In this paper we analyze the Diffusion Prior model that generates CLIP image embedding conditioned on CLIP text embedding and its applicability for text gudied image editing in detail. We discover that the Diffusion Prior model can be used to perform text guided conceptual editing in the CLIP latent space as described in Sec.\ref{sec:conceptual}. We combine this with structural editing (ref. Sec.\ref{sec:structural}) using existing methods adapted to the CLIP image embedding conditioned Diffusion Decoder model that we train to perform text guided image editing as described in Sec.\ref{sec:preditor}. We show qualitatively that the results are comparable or better than ablated versions in Sec.\ref{sec:ablation} and state of the art approaches in Sec.\ref{sec:related}. Discussion on the importance of the conceptual edit step and its visualizations are provided in Sec.\ref{sec:discussion} followed by optional mask guidance for localized edits and effect of hyperparameters for controllable editing in Sec.\ref{sec:discussion}.

In the Appendix, we describe the detailed architecture of HDM in Sec.\ \ref{appendix:training}. Examples of text-to-image generation and image variations using our HDM, that combines the LAION prior with the trained LDM are given in Sec.\ \ref{appendix:hdm}. We emphasize that examples for text to image generation and image variations are provided to show that the LDM model we trained combined with the LAION prior works well. The focus of the paper is however on text guided image editing using this HDM architecture. We discuss intuitions of the proposed refinement setup in Sec.\ \ref{appendix:training}. We emphasize that DDIM inversions on CLIP image embedding conditioned Diffusion Decoder is relatively more effective than text conditioned models in Sec.\ \ref{appendix:inversion}. To better understand the effect of \emph{conceptual edit} step in the CLIP latent space, we try to intuitively illustrate the process in Sec.\ref{appendix:shifted} and compare it with shifted diffusion. We further visualize the intermediate outputs and perform CLIP score analysis at various injection points $c$ to gain further intuition in Sec.\ \ref{appendix:prior_visual}. We notice that the edited embedding generated by the Diffusion Prior after \emph{conceptual edit} step qualitatively shows closeness to base image and the edit text. We also quantitatively measure the relevance of generated edits to the edit prompt and structural fidelity to the base image in Sec.\ref{appendix:quant} and compare with baselines. More results with and without masked guidance and for a variety of base images and edit prompts are shown in Sec.\ \ref{appendix:examples} followed by a discussion on the limitations in Sec.\ref{appendix:limitations}.
\section{Training HDM}
\label{appendix:training}
We describe the Hybrid Diffusion Model consisting of a Diffusion Prior and a Diffusion Decoder. 
\subsection{Diffusion Prior Model}
\label{appendix:proposed-prior}
The Prior model is a denoising diffusion model as proposed in \cite{dalle2}. that generates a normalized CLIP L/14 image embedding $z_x$ conditioned on an input prompt $y$. The Diffusion Prior $\mathcal{P}_\theta(z_x | y)$ parameterized by $\theta$ is a Causal Transformer \cite{dalle2} that takes as input a random noise $\epsilon$ sampled from $\mathcal{N}(0, I)$ and a CLIP text embedding $z_y$ = $[z_t, w_{1}, w_{2}, ..., w_n]$ where $z_t$ is the l2 normalized text embedding while $w_{i}$ is the per token encoding, both from a pretrained CLIP L/14 text encoder \cite{clip}. The maximum sequence length for CLIP is $77$ and hence $z_{y}$ has dimensions $78$ x $768$. The LAION prior\footnote{https://huggingface.co/nousr/conditioned-prior/blob/main/vit-l-14/prior\_aes\_finetune.pth} was trained using the code \footnote{https://github.com/lucidrains/DALLE2-pytorch} on LAION\footnote{https://laion.ai} data with ground truth $z_y$ and $z_x$ from text-image $(y, x)$ pairs using the same setup and MSE objective as denoising diffusion models \cite{ddpm, dalle2}. We use the pretrained LAION prior as is for inference.
\subsection{Diffusion Decoder Model}
\label{appendix:proposed-ldm}
For the Diffusion Decoder, instead of the complex multi-stage pixel diffusion model proposed in \cite{dalle2}, we train a custom latent space model inspired by \cite{ldm} for memory and compute efficiency. Our LDM is a denoising diffusion model $\mathcal{D}_\phi(x | z_x)$ parameterized by $\phi$ that takes as input random sample $\epsilon$ from $\mathcal{N}(0, I)$ and the CLIP image embedding $z_x$ to generate the VAE \cite{ldm} latent $\mathbf{z}$. It is to be noted that the bold notation for VAE latent $\mathbf{z}$ is different from the notation for CLIP embeddings $z$. The generated latent $\mathbf{z}$ is passed through a frozen decoder VAE$_{dec}$ of a pretrained VAE to generate the final image $\mathbf{x}$. The LDM's architecture is unchanged from \cite{ldm} except the conditioning input to the UNet \cite{unet} is modified to be normalized CLIP L/14 embedding $z_x$ instead of $z_y$ as in Stable Diffusion. During inference, we can provide $z_x$ for an image $x$ to generate variations of it \cite{dalle2} or provide the generated embedding $\hat{z}_x$ from the Diffusion Prior to generate image from text $\mathbf{y}$ for text-to-image generation. We base our code on \cite{ldm}\footnote{https://github.com/CompVis/latent-diffusion} and change only the input condition.

To summarize, we follow the same two step hybrid architecture for text-to-image generation as \cite{dalle2}. For the prior, we use the pre-trained and publicly available LAION model and for the decoder, we train a custom LDM conditioned on normalized CLIP L/14 image embeddings. The Diffusion Prior, pre-trained VAE for the LDM and CLIP L/14 models are all frozen while the Diffusion Decoder is trained. 
\begin{figure*}[t]
 \centering
 \begin{tabular}{c@{\hspace{0.3cm}}c@{\hspace{0.3cm}}c@{\hspace{0.3cm}}}
 \centering
 \parbox{.3\linewidth}{A cute puppy on the beach}&\parbox{.3\linewidth}{A modern glass house in a rainforest, by leonid afremov}&\parbox{.3\linewidth}{an illustration of a dog wearing a graduation hat}\\ \\
 \includegraphics[width=.3\linewidth]{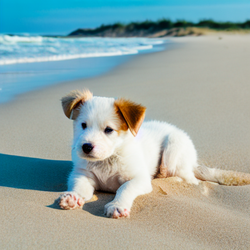}&
 \includegraphics[width=.3\linewidth]{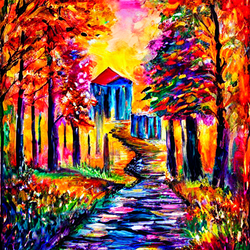}&
 \includegraphics[width=.3\linewidth]{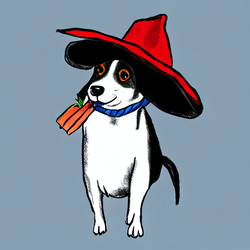} \\ \\
 \parbox{.3\linewidth}{an ultra realistic rendering of a futuristic sci-fi spaceship corridor, artstation}&\parbox{.3\linewidth}{highly detailed landscape of a snowy field under clear blue sky with sunbeams}&\parbox{.3\linewidth}{modern bathroom with black faucet and shower on wall}\\ \\
 \includegraphics[width=.3\linewidth]{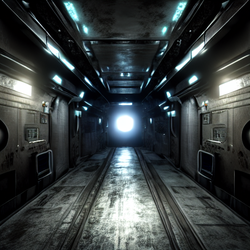}&
 \includegraphics[width=.3\linewidth]{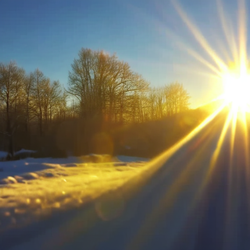}&
 \includegraphics[width=.3\linewidth]{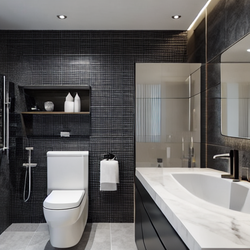} \\ \\
 \parbox{.3\linewidth}{Vintage house in a thunderstorm at night}&\parbox{.3\linewidth}{winter forest background with white snow, pixelart}&\parbox{.3\linewidth}{an image of alaska in the night sky with aurora and milkyway}\\ \\
 \includegraphics[width=.3\linewidth]{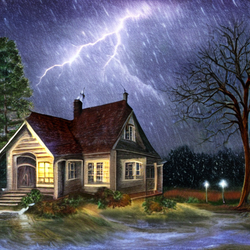}&
 \includegraphics[width=.3\linewidth]{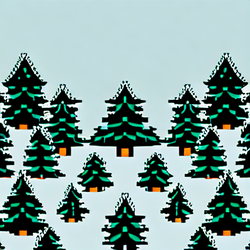}&
 \includegraphics[width=.3\linewidth]{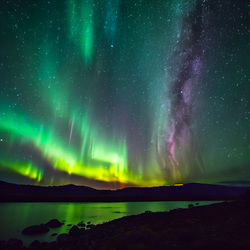}
 \end{tabular}
 \caption{Example text-to-image generations from the HDM. Note that even though the LDM is trained on Stock data without information about modifiers and styles, plugging the LAION prior that generates CLIP embeddings helps the LDM generate reasonably relevant content.}
 \label{fig:appendix-hdm}
\end{figure*}

\begin{figure*}[t]
 \centering
 \begin{tabular}{c@{\hspace{0.2cm}}c@{\hspace{0.2cm}}c@{\hspace{0.2cm}}c@{\hspace{0.2cm}}}
 \centering
 Input & \multicolumn{3}{c}{Variations}\\
 \includegraphics[width=.23\linewidth]{images/appendix_hdm/vintage-house-in-a-thunderstorm-at-night.png}&
 \includegraphics[width=.23\linewidth]{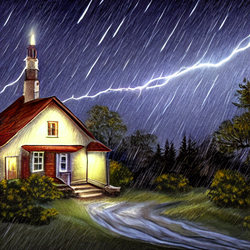}&
 \includegraphics[width=.23\linewidth]{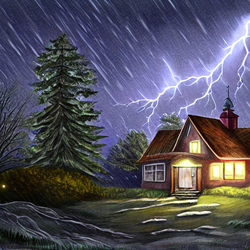}&
 \includegraphics[width=.23\linewidth]{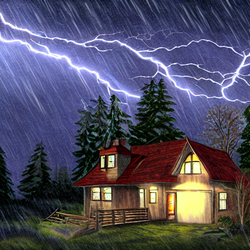} \\
 &
 \includegraphics[width=.23\linewidth]{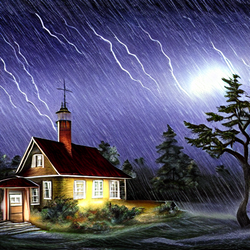}&
 \includegraphics[width=.23\linewidth]{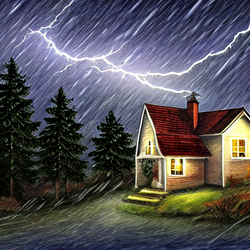}&
 \includegraphics[width=.23\linewidth]{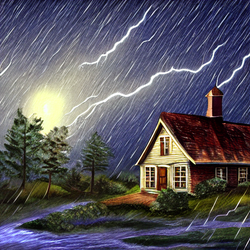}\\ \\
 \includegraphics[width=.23\linewidth]{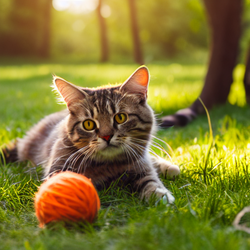}&
 \includegraphics[width=.23\linewidth]{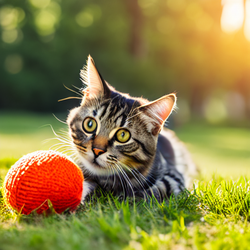}&
 \includegraphics[width=.23\linewidth]{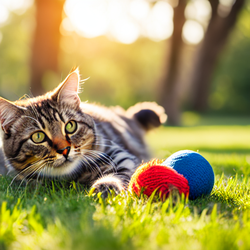}&
 \includegraphics[width=.23\linewidth]{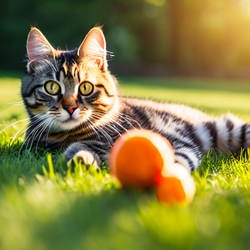} \\
 &
 \includegraphics[width=.23\linewidth]{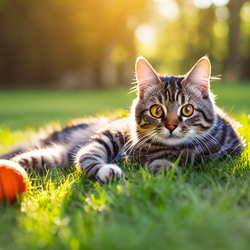}&
 \includegraphics[width=.23\linewidth]{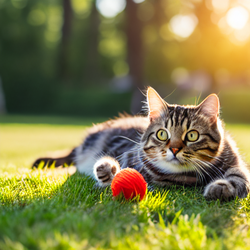}&
 \includegraphics[width=.23\linewidth]{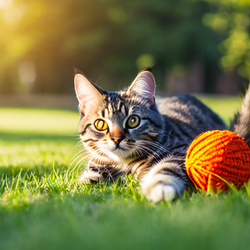}
 \end{tabular}
 \caption{Example image variations generated by conditioning the Diffusion Decoder on the CLIP image embedding of the input image.}
 \label{fig:appendix-variations}
\end{figure*}
\subsection{Text-to-Image Generation with HDM}
\label{appendix:hdm}
The LDM model is trained on an internal %
77M images dataset after filtering for humans and text to simplify the diversity in concepts. As a result, our HDM cannot create human faces or text in the images. Compared to baselines mentioned in Sec.\ \ref{sec:baselines} that use Stable Diffusion or Imagen models that were trained on large scale data, ours is significantly smaller. However, the model generates high quality images within the training data distribution. Some examples of text-to-image generation using the LAION prior and the custom LDM are shown in Fig.\ \ref{fig:appendix-hdm}. We can see that the HDM has wide concept coverage. Even though the LDM has not seen artist styles (e.g. Leonid Afremov) or modifiers, since the LAION prior trained on such data can generate CLIP embeddings that the LDM is conditioned on, it is able to adapt to novel concepts. However, this adaptation does have limits and the LDM cannot create completely unseen concepts such as human faces.
\begin{figure*}[t]
 \centering
 \begin{tabular}{c@{\hspace{0.2cm}}|c@{\hspace{0.2cm}}c@{\hspace{0.2cm}}c@{\hspace{0.2cm}}c@{\hspace{0.2cm}}}
 \centering
 Real Image & SD 1.0 & HDM 1.0 & SD 0.7 & HDM 0.7\\ 
 \includegraphics[width=.18\linewidth]{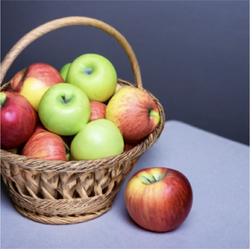}&
 \includegraphics[width=.18\linewidth]{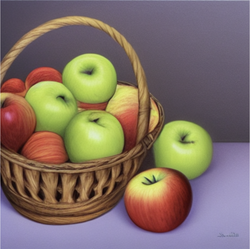}&
 \includegraphics[width=.18\linewidth]{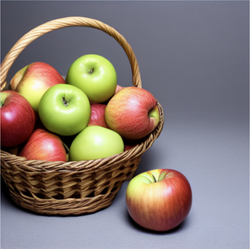}&
 \includegraphics[width=.18\linewidth]{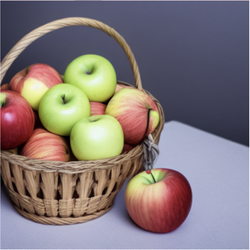}&
 \includegraphics[width=.18\linewidth]{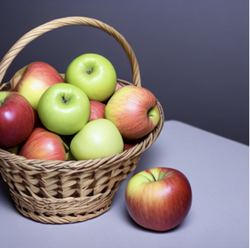} \\
 \includegraphics[width=.18\linewidth]{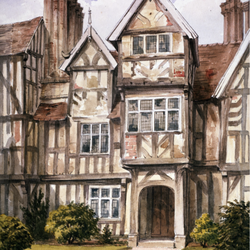}&
 \includegraphics[width=.18\linewidth]{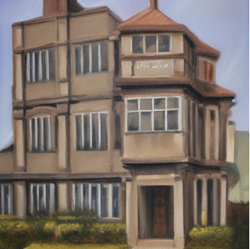}&
 \includegraphics[width=.18\linewidth]{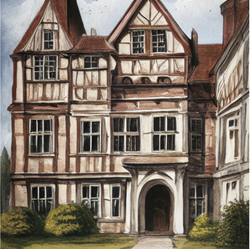}&
 \includegraphics[width=.18\linewidth]{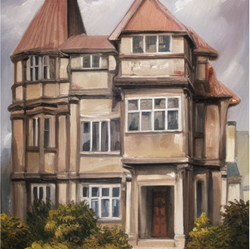}&
 \includegraphics[width=.18\linewidth]{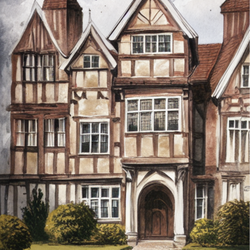} \\
 \includegraphics[width=.18\linewidth]{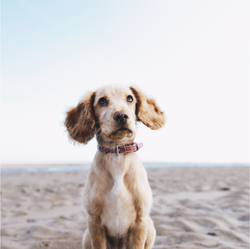}&
 \includegraphics[width=.18\linewidth]{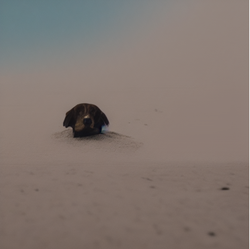}&
 \includegraphics[width=.18\linewidth]{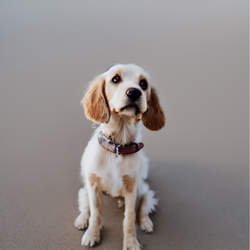}&
 \includegraphics[width=.18\linewidth]{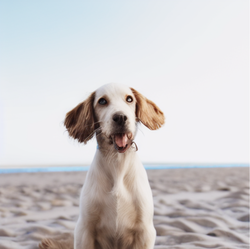}&
 \includegraphics[width=.18\linewidth]{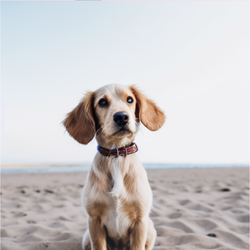} 
 \end{tabular}
 \caption{Example real image inversions using stable diffusion and the diffusion decoder in HDM using reverse DDIM. Suggested to view the images zoomed in for clarity. Text used for inversion with stable diffusion given in text. We can clearly see the advantages of using CLIP image embedding conditioning in the decoder for inversions.}
 \label{fig:appendix-inversion}
\end{figure*}
Moreover, we can also see from Fig.\ \ref{fig:appendix-variations} that the LDM can generate reasonable and relevant variations of an image when conditioned on the CLIP embedding of an input image. This highlights the amount of information captured in the CLIP embedding as also observed in \cite{dalle2}. %

\section{Inversion in HDM}
\label{appendix:inversion}
One of the major advantages of our HDM architecture inspired by DALLE-2 is that the diffusion decoder is conditioned on CLIP image embedding. As a result, to invert a real image to get a deterministic noise map that can be used for reconstruction, we can condition on the CLIP embedding of the image to be inverted. On the other hand, models such as Stable Diffusion and Imagen are conditioned directly on text. Inverting a real image to get the random noise using such models requires a manually defined text prompt. However, not all text prompts with the real image are invertible. It has been observed previously \cite{nullinversion, edict} that the more detailed and precise the prompt is, the better the inversion and reconstruction quality. Consequently, these works have proposed novel optimization based techniques to make inversions better \cite{nullinversion,edict} with Stable Diffusion. 

We contend that the CLIP embedding of an image has more information about the image than a manually defined text prompt. Since variations work well with the LDM model, we also test if inversions are better. Fig.\ \ref{fig:appendix-inversion} shows example inversion with Stable Diffusion as well as the Diffusion Decoder from our HDM. For Stable Diffusion we used the v1.3 checkpoint with the text prompts \emph{`Basket of apples'}, \emph{`An oil painting of a building'} and \emph{`A white and brown dog sitting on the beach'} corresponding to images in top down order as shown in Fig.\ \ref{fig:appendix-inversion}. We used default settings for all parameters during sampling. We also tried using different versions of text and have shown the results for the best setting. For example, for the third image in Fig.\ \ref{fig:appendix-inversion}, we tried \emph{`A close-up photo of a white dog sitting on the beach, looking up at the camera'}, \emph{`A realistic photo of a white dog with red collar sitting on the beach'} and others. The values 1.0 and 0.7 correspond to the strength of reverse DDIM. If the max number of sampling steps is 100, then 1.0 corresponds to running all the 100 steps of DDIM inversion to get to $\mathbf{z_{1000}}$ whereas 0.7 corresponds to stopping approximately at $\mathbf{z_{700}}$. To get the reconstructions, we do regular DDIM sampling conditioned on the input text (CLIP image embedding in HDM) using the inverted noise. We can clearly see that CLIP image embedding conditioned decoder can do better inversions. Even though going all the way to $z_T$ does not reconstruct input image accurately even with HDM, it is relatively much closer and reaches accurate inversions with 70\% of the steps. Keeping up with observations in \cite{nullinversion, edict}, we perform reverse DDIM with cfg$= 1.0$ but use a default of cfg$= 5.0$ for regular sampling to generate the images. Performing reverse DDIM for cfg$\geq 1.0$ causes noisy and irregular results but still performs better than doing the same with text conditioned models as shown in Fig.\ \ref{fig:appendix-inversion}. We believe this is another important reason besides \emph{conceptual editing} with the diffusion prior that helps PRedItOR achieve structure preserving edits without any additional objectives or optimizations. 

\section{Intuition for Prior Refinement}
\label{appendix:shifted}
A recently proposed technique called Shifted Diffusion \cite{shifted_diffusion} is one of the few papers to look into the Diffusion Prior architecture. They propose an alternative formulation for the prior to sample from a random CLIP image embedding (sampled from a collection of image clusters) instead of random Gaussian noise. The authors hypothesize that the space spanned by valid CLIP image embeddings is small and navigating from random noise to this space is sub-optimal and causes issues with relevance. 
\begin{figure}[t]
 \centering
 \includegraphics[width=0.5\linewidth]{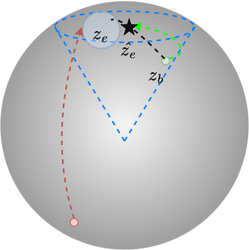}
 \caption{An intuitive illustration of the \emph{conceptual editing} by PRedItOR compared with a regular sampling step.}
 \label{fig:intuition}
\end{figure}
\begin{figure*}[t]
 \centering
 \begin{tikzpicture}
 \draw [->] (0,0) -- (10,0) node [midway, above] {$t$};
\end{tikzpicture} \\ \vspace{0.05in} 
 \rotatebox[x=1mm,y=1mm]{90}{\parbox{1.5cm}{\emph{basket of kittens}}}\hspace{0.05in}
 \includegraphics[width=.1\linewidth]{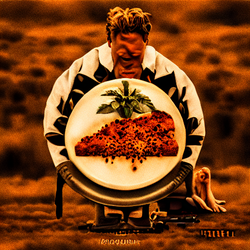} 
 \includegraphics[width=.1\linewidth]{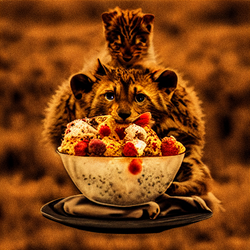}
 \includegraphics[width=.1\linewidth]{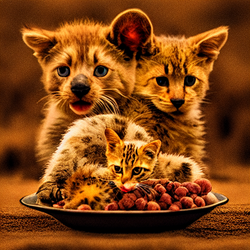}
 \includegraphics[width=.1\linewidth]{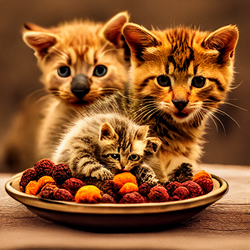}
 \includegraphics[width=.1\linewidth]{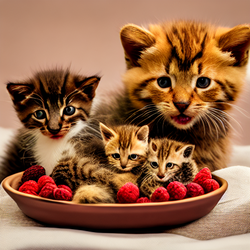}
 \includegraphics[width=.1\linewidth]{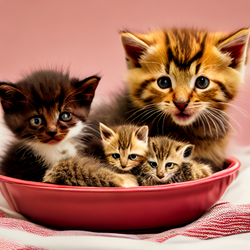}
 \includegraphics[width=.1\linewidth]{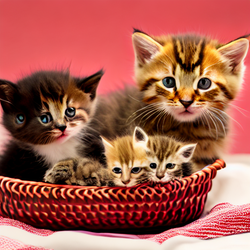}
 \includegraphics[width=.1\linewidth]{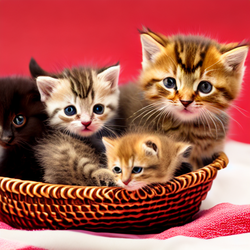}
 \includegraphics[width=.1\linewidth]{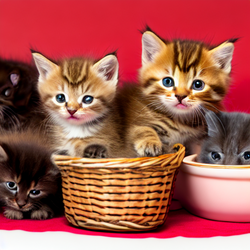}
 \\ \vspace{0.05in} 
 \rotatebox[x=1mm,y=4mm]{90}{\parbox{1.5cm}{\emph{flooded}}}\hspace{0.2in}
 \includegraphics[width=.1\linewidth]{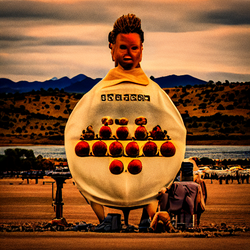} 
 \includegraphics[width=.1\linewidth]{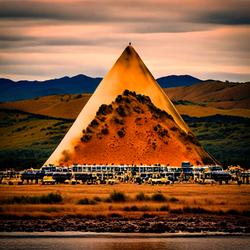}
 \includegraphics[width=.1\linewidth]{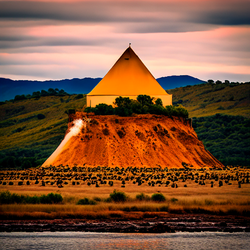}
 \includegraphics[width=.1\linewidth]{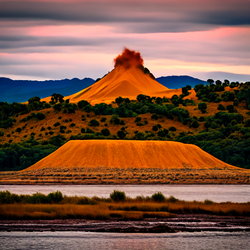}
 \includegraphics[width=.1\linewidth]{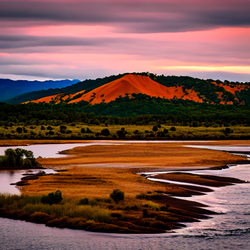}
 \includegraphics[width=.1\linewidth]{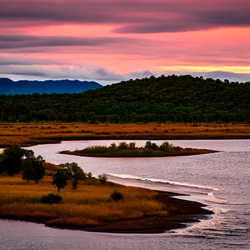}
 \includegraphics[width=.1\linewidth]{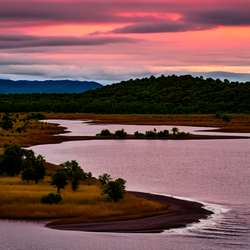}
 \includegraphics[width=.1\linewidth]{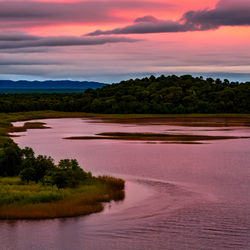}
 \includegraphics[width=.1\linewidth]{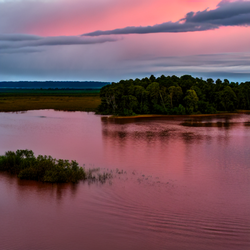}
 \\ \vspace{0.05in} 
 \rotatebox[x=1mm,y=1mm]{90}{\parbox{1.5cm}{\emph{an oil painting}}}\hspace{0.05in}
 \includegraphics[width=.1\linewidth]{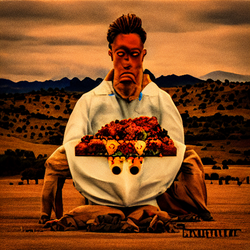} 
 \includegraphics[width=.1\linewidth]{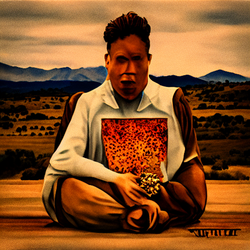}
 \includegraphics[width=.1\linewidth]{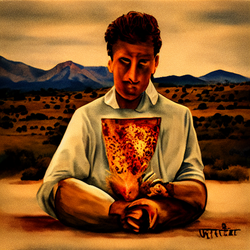}
 \includegraphics[width=.1\linewidth]{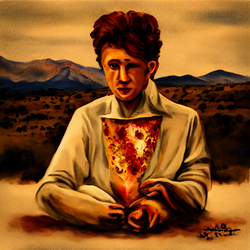}
 \includegraphics[width=.1\linewidth]{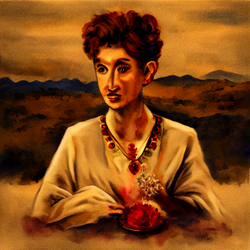}
 \includegraphics[width=.1\linewidth]{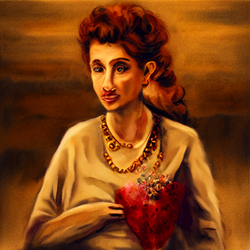}
 \includegraphics[width=.1\linewidth]{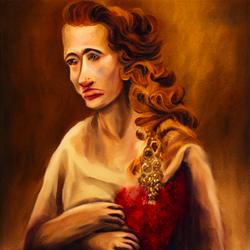}
 \includegraphics[width=.1\linewidth]{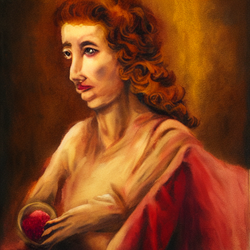}
 \includegraphics[width=.1\linewidth]{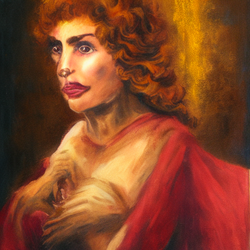}
 \caption{Visualizing the intermediate predictions from the prior as timestep $t$ decreases by passing it through the decoder after normalization since the intermediate outputs from prior are not normalized. Values of $t$ correspond to 891, 792, 693, 594, 495, 396, 297, 198, 0 from left to right.}
 \label{fig:appendix_conceptual_intermediate}
\end{figure*}
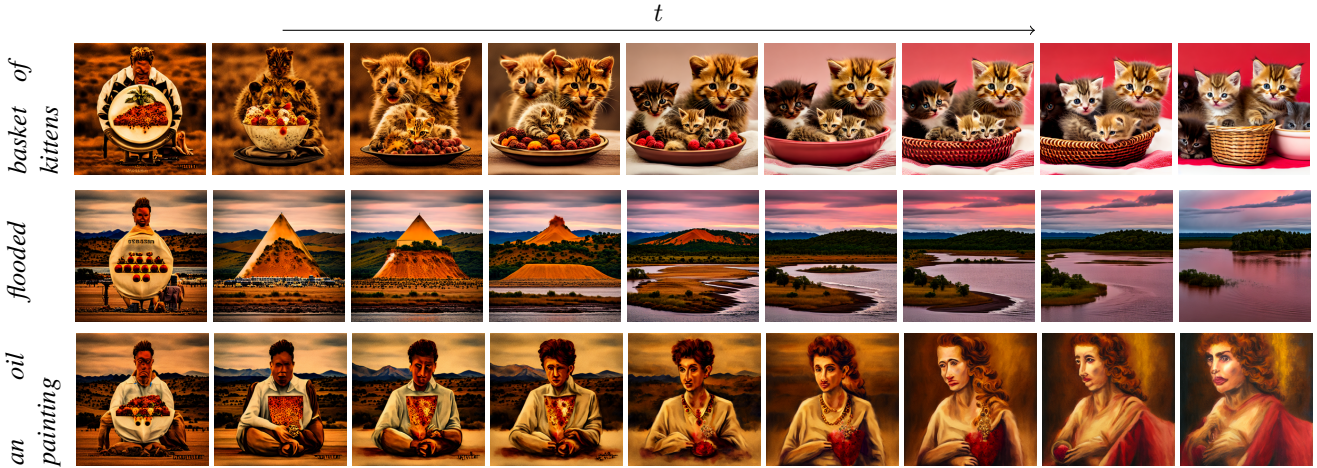
\begin{figure*}[t]
 \centering
 \begin{tikzpicture}
 \draw [->] (0,0) -- (10,0) node [midway, above] {$c$};
\end{tikzpicture} \\ \vspace{0.05in} 
 \rotatebox[x=1mm,y=1mm]{90}{\parbox{1.5cm}{\emph{basket of kittens}}}\hspace{0.05in}
 \includegraphics[width=.1\linewidth]{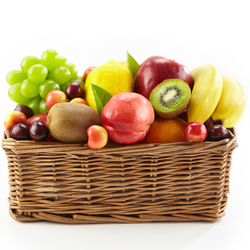} 
 \includegraphics[width=.1\linewidth]{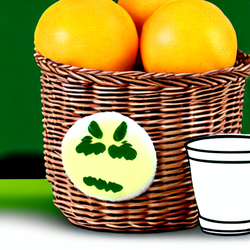}
 \includegraphics[width=.1\linewidth]{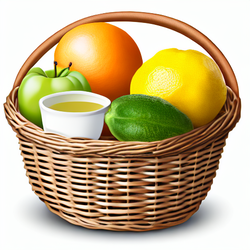}
 \includegraphics[width=.1\linewidth]{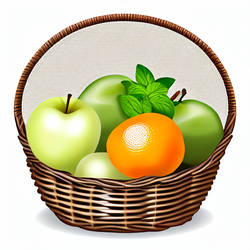}
 \includegraphics[width=.1\linewidth]{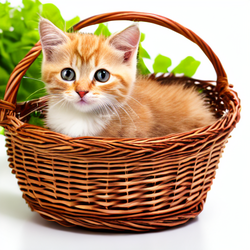}
 \includegraphics[width=.1\linewidth]{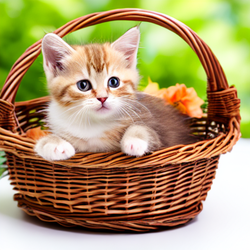}
 \includegraphics[width=.1\linewidth]{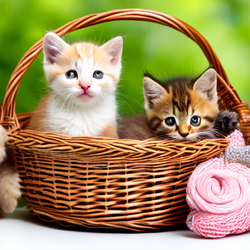}
 \includegraphics[width=.1\linewidth]{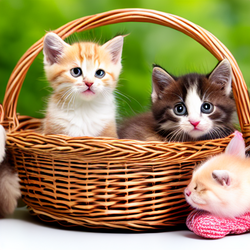}
 \includegraphics[width=.1\linewidth]{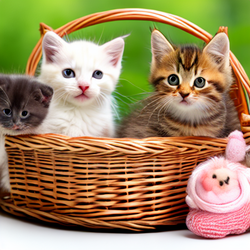}
 \\ \vspace{0.05in} 
 \rotatebox[x=1mm,y=4mm]{90}{\parbox{1.5cm}{\emph{flooded}}}\hspace{0.15in}
 \includegraphics[width=.1\linewidth]{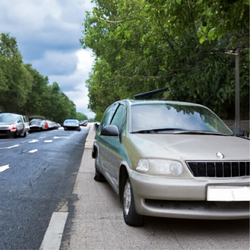} 
 \includegraphics[width=.1\linewidth]{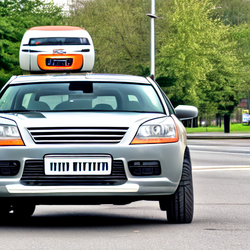}
 \includegraphics[width=.1\linewidth]{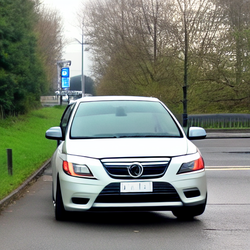}
 \includegraphics[width=.1\linewidth]{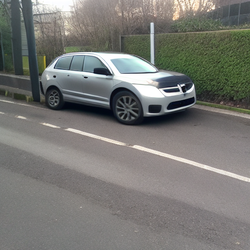}
 \includegraphics[width=.1\linewidth]{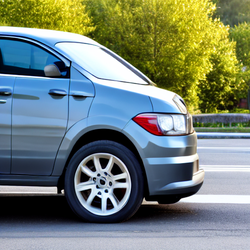}
 \includegraphics[width=.1\linewidth]{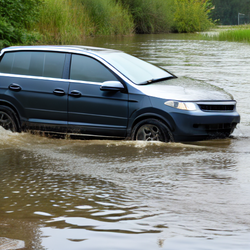}
 \includegraphics[width=.1\linewidth]{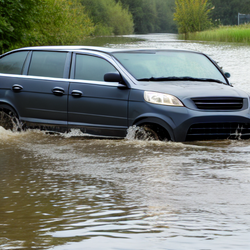}
 \includegraphics[width=.1\linewidth]{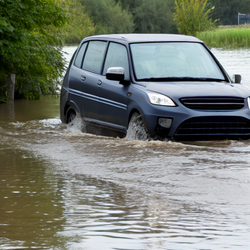}
 \includegraphics[width=.1\linewidth]{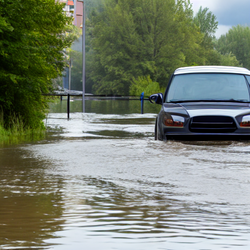}
 \\ \vspace{0.05in} 
 \rotatebox[x=1mm,y=1mm]{90}{\parbox{1.5cm}{\emph{an oil painting}}}
 \includegraphics[width=.1\linewidth]{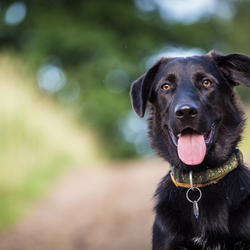} 
 \includegraphics[width=.1\linewidth]{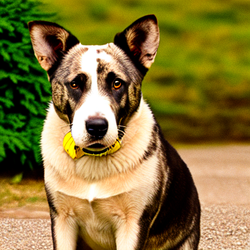}
 \includegraphics[width=.1\linewidth]{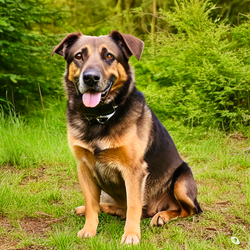}
 \includegraphics[width=.1\linewidth]{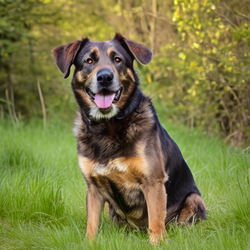}
 \includegraphics[width=.1\linewidth]{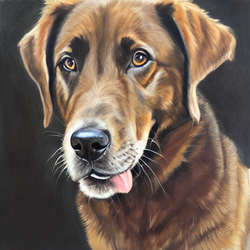}
 \includegraphics[width=.1\linewidth]{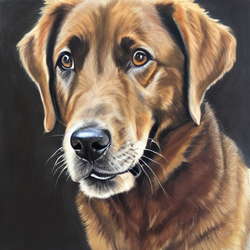}
 \includegraphics[width=.1\linewidth]{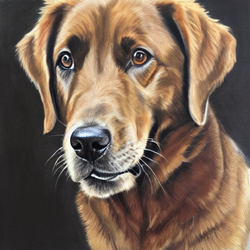}
 \includegraphics[width=.1\linewidth]{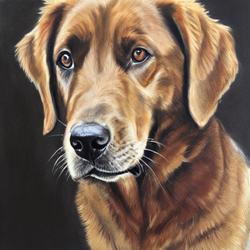}
 \includegraphics[width=.1\linewidth]{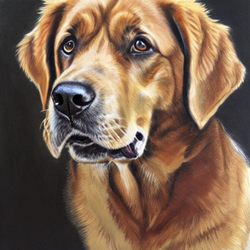}
 \caption{Visualizing the embedding generated by the \emph{conceptual edit} step for three samples as $c$ increases. Values of $c$ correspond to 0.2, 0.35, 0.45, 0.55, 0.58, 0.6, 0.65, 0.8 from left to right. When $c$ is 0.2, 20 DDIM steps of denoising is performed starting from timestep $t=200$ to $t=0$.}
 \label{fig:appendix_conceptual_refinement}
\end{figure*}
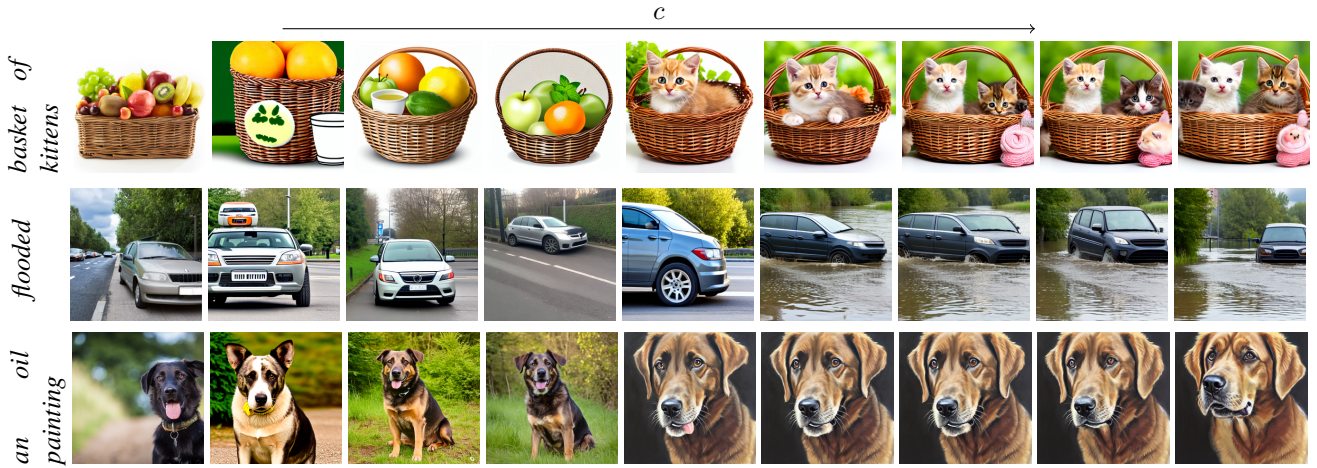
We try to intuitively understand how the \emph{conceptual edit} step proposed in Sec.\ \ref{sec:conceptual} works. An illustration is shown in Fig.\ \ref{fig:intuition} inspired by \cite{shifted_diffusion}. In our setup, we inject a valid CLIP image embedding $z_b$ of the base image at some intermediate timestep $t$ in the diffusion prior. The green dashed arrow from $z_b$ to $\hat{z_e}$ corresponds to a possible depiction of the path the prior takes during the \emph{conceptual edit} step, when conditioned on edit text $y_e$. The red dashed line corresponds to a possible path from random noise during regular sampling. Based on the injection timestep $t_c$, we hypothesize that the embedding generated by the prior $\hat{z_e}$ lies somewhere on the interpolation path between $z_b$ and $z_e$ resulting in capturing both the concepts effectively. As $t_c$ increases, $\hat{z_e}$ gets closer to the concept space surrounding $z_e$ corresponding to $y_e$. The \emph{conceptual edit} step can be considered as imitating the shifted diffusion process resulting in generations that visually depict a latent walk as shown in Fig.\ \ref{fig:conceptual}. The images visualized in Fig.\ \ref{fig:conceptual} are then points in the black dashed curve joining $z_b$ and $z_e$. The blue cone depicts the space spanned by all valid CLIP image embeddings \cite{shifted_diffusion}. It is to be noted that the LAION \emph{Diffusion Prior} used in PRedItOR works on normalized CLIP embedding and that would ideally translate to a unit circle rather than a cone within the hypersphere but the visualization is still applicable. We show further analysis in Sec.\ \ref{appendix:prior_visual}. We emphasize that there is more to understand about the diffusion process in a low dimensional latent space and further research could generate interesting new models and applications.
\begin{figure*}[ht]
 \centering
 \includegraphics[width=0.33\linewidth]{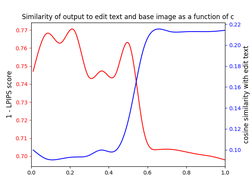}
 \includegraphics[width=0.33\linewidth]{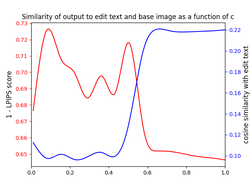}
 \includegraphics[width=0.33\linewidth]{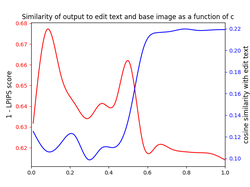}
 \caption{(1 - LPIPS) score measures fidelity between generated image and base image whereas CLIP measures the generated image's relevance to edit text for various values of $c$. The three plots are plotted for \emph{structural edit} strengths $s$ 0.4, 0.5 and 0.5 in left to right order.}
 \label{fig:appendix_lpips}
\end{figure*}
\section{Understanding PRedItOR with Visualizations}
\label{appendix:prior_visual}
Previous research \cite{ddpm, ddim} has visualized the intermediate outputs of a DM and observe the gradual addition and removal of visual noise. Though few works have utilized the Diffusion Prior formulation \cite{dalle2, make_a_video, runway, shifted_diffusion}, there is no deeper understanding of the diffusion process in the CLIP latent space. To further understand the workings of the Diffusion Prior, we visualize the intermediate outputs from the prior during regular sampling and \emph{conceptual edit} refinement for a given text prompt, starting from the same random seed for all prompts. In Fig.\ \ref{fig:appendix_conceptual_intermediate}, we can see that as the denoising timestep tends to 0, the concept corresponding to the input text prompt gets better represented in the generated image. We can think of these intermediate images as being generated from intermediate points along the red dashed path in Fig.\ \ref{fig:intuition}. We can also observe that images generated from intermediate outputs at the initial timesteps do not have any visually recognizable concept. This could be because the outputs have not reached within the bounds of the valid CLIP embedding space (highlighted by blue dashed cone in Fig.\ \ref{fig:intuition}).
\begin{figure}[h]
 \centering
 \includegraphics[width=0.7\linewidth]{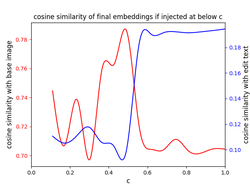}
 \caption{CLIP score of base image and edit text with the generated image at different injection points $c$.}
 \label{fig:appendix_clip}
\end{figure}
In Fig.\ \ref{fig:appendix_conceptual_refinement} we visualize the final embedding generated by the \emph{conceptual edit} step for various values of $c$. When $c$ is lower, the prior gets fewer steps to refine the embedding corresponding to input text resulting in little to no information from the input text. As $c$ increases however, the prior runs enough text guided \emph{conceptual denoising} steps on the injected embedding. For values between 0.55 and 0.8 we can see that the generated images have a good balance of the input image as well as the edit text. In the second row in Fig.\ \ref{fig:appendix_conceptual_refinement}, the concept from edit text 'flooded' is combined with the concept of 'car' from the input image as $c$ increases. We can use one of these embeddings for the subsequent \emph{structural edit} step to complete the editing. 

We plot CLIP relevance to edit text and (1 - LPIPS) fidelity \cite{lpipspaper} to base image as a function of the \emph{conceptual edit} strength $c$ in Fig.\ \ref{fig:appendix_lpips} following \cite{imagic, edict}. The three plots correspond to \emph{structural edit} strength $s$ 0.4, 0.5 and 0.6. These are averaged over the three examples shown in Fig.\ \ref{fig:appendix_conceptual_refinement}. We can see that value between 0.4 and 0.7 is the sweet spot for optimal relevance to both edit text and base image. We can also observe that as $s$ increases (the third plot), fidelity to base image starts to reduce around the highlighted region. Both $c$ and $s$ can be controlled by the user to generate varied results.

To further understand the effect of \emph{conceptual edit} step on the generation and to quantify the observation in Fig.\ \ref{fig:appendix_conceptual_refinement}, we plot the CLIP score between generated image and base image as well as generated image and edit text as a function of the \emph{conceptual edit} strength $c$ similar to \cite{instructpix2pix}. The plot as shown in Fig.\ \ref{fig:appendix_clip} is the average over the three examples shown in Fig.\ \ref{fig:appendix_conceptual_refinement} with increments of 0.1 for $c$. We can see that as $c$ increases, fidelity to base image decreases while to that of the edit text increases. The transition is smoother between 0.5 and 0.7. We also notice some fluctuations at lower values of $c$. We believe this is because the prior does not have enough time to get to the right direction and embedding which is also the reason why the third column of images in Fig.\ \ref{fig:appendix_conceptual_refinement} corresponding to $c=0.2$ do not have reasonable quality.

\section{Quantitative Evaluation}
\label{appendix:quant}
We emphasize that there is no publicly available dataset for large scale quantitative evaluation and there are no reliable metrics to measure editing performance. Even though some existing metrics are used for evaluation in related works such as \cite{plugnplay}, there are a number of hyperparameters to control in most editing methods and the generations can be diverse with varied subjective preference. However, we still perform a small scale quantitative evaluation to keep with existing works. 

\begin{table}[h]
 \centering
 \begin{tabular}{p{0.35\linewidth}p{0.15\linewidth}p{0.15\linewidth}p{0.15\linewidth}}
 \toprule
 Baseline & CLIP$\uparrow$& LPIPS$\downarrow$ & St.Dist.$\downarrow$\\
 \midrule 
P2P\cite{prompt2prompt} & 0.197 & \textbf{0.273} & 0.024* \\
P\&P\cite{plugnplay} & 0.221 & 0.474 & 0.054 \\
SD\footnotesize{(S $0.70$)}\cite{ldm} & 0.224* & 0.552 & 0.073 \\
SD\footnotesize{(r $0.70$)}\cite{ldm} & 0.224* & 0.369 & 0.048 \\
\midrule
Ours\footnotesize{($c$=0.55, $s$=0.35)} & 0.177 & 0.281* & \textbf{0.023} \\
Ours\footnotesize{($c$=0.55, $s$=0.70)} & 0.192 & 0.480 & 0.079 \\
Ours\footnotesize{($c$=0.75, $s$=0.35)} & 0.213 & 0.319 & 0.030 \\
Ours\footnotesize{($c$=0.75, $s$=0.70)} & \textbf{0.234} & 0.558 & 0.103 \\
 \bottomrule
 \end{tabular}
 \caption{Quantitative comparison between proposed PRedItOR and baseline methods for editing, averaged over 100 samples. S and r correspond to SDEdit and rDDIM for structural edit. The float value next to the alphabets is the strength used. Best values are bolded while the second best values are highlighted with a *.}
 \label{tab:quant}
 \vspace{-0.1in}
\end{table}

We first consider 4 classes of images \emph{dog}, \emph{cake}, \emph{horse}, \emph{flowers} corresponding to the most common class of base images used in most editing works. We get 5 random images per class by searching with the class name from UnSplash\footnote{https://unsplash.com}. This gives us 20 base images. We then manually provide 5 edit prompts per class of images to get a total of 100 edited images from our method as well as baselines. For baselines we choose \textbf{SD(SDEdit)} \cite{ldm}, \textbf{SD(rDDIM)} \cite{ldm}, \textbf{P2P} \cite{prompt2prompt} and \textbf{PnP} \cite{plugnplay}. We generate edits using default parameters for P2P and PnP and use 2 different parameters 0.35 and 0.7 as strength for SDEdit and rDDIM for SD. We use the same 0.35 and 0.7 for $s$ in our method for fair comparison. We use the publicly available implementation for P2P and P\&P. For P2P since it does not directly work on real images, we use it with \cite{nullinversion} with all default parameters. For the prompts we use \emph{'A photo of a \{class\}'} as the base prompt and \emph{'A photo of a \{edit\}'} or  \emph{'A \{edit\} of a \{class\}'} as the edit prompt in P2P. For example, \emph{'A photo of a dog'} and \emph{'A photo of a dog stuck in flood'} or \emph{'A oil painting of a dog'}. For P\&P we use the default edit prompts and parameters except the number of steps for reverse DDIM. The authors suggest $999$ i.e the full inversion but we use $50$ corresponding to the number of sampling steps to be fair with all other methods that use the same parameter. The authors also use $50$ for generated images that are easier to invert. Given these constraints, we keep the number as $50$ for quantitative evaluation.

\begin{figure*}[t]
 \centering
 \begin{tabular}{c@{\hspace{0.15cm}}|c@{\hspace{0.15cm}}c@{\hspace{0.15cm}}c@{\hspace{0.15cm}}c@{\hspace{0.15cm}}c@{\hspace{0.15cm}}}
 \centering
 &PRedItOR&P2P&P\&P&SD(SDEdit)&SD(rDDIM)\\
 \hline
 \emph{children's drawing}&&&&&\\
 \includegraphics[height=1in, width=1in]{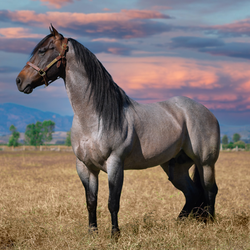}&
 \includegraphics[height=1in, width=1in]{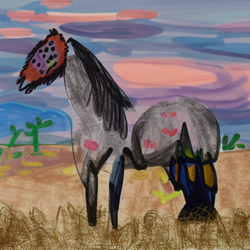}&
 \includegraphics[height=1in, width=1in]{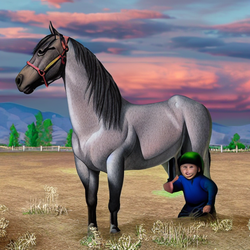}&
 \includegraphics[height=1in, width=1in]{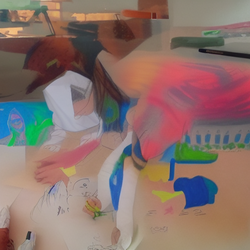}&
 \includegraphics[height=1in, width=1in]{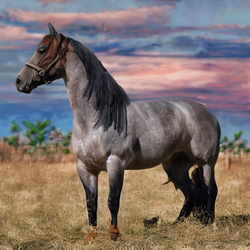}&
 \includegraphics[height=1in, width=1in]{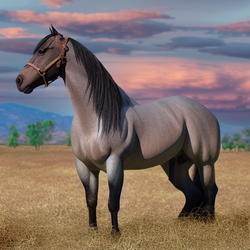} \\
 LPIPS$\downarrow$&0.441&0.302&0.647&0.372&0.206\\
 St.Dist.$\downarrow$&0.025&0.018&0.104&0.012&0.004\\
 CLIP$\uparrow$&0.221&0.204&0.229&0.124&0.155\\
 \end{tabular}
    \caption{Example edits from baselines and PRedItOR along with quantitative metrics. Though PRedItOR shows worser scores (lower is better) for structural fidelity metrics LPIPS and St.Dist., compared to baselines such as P2P and SD, we can see visually that PRedItOR performs the desired edit more effectively while the baselines do not . Similarly, P\&P scores highest for CLIP relevance but has compltely lost all information about the base image.}
    \label{fig:appendix_unfair}
    \vspace{-0.1in}
\end{figure*}
For metrics we use CLIP \cite{clip} score between generated image and input edit prompt to measure relevance, LPIPS \cite{lpipspaper}  and StructSim. \cite{plugnplay} to measure structural fildelity between edited image and base image. St.Dist. is structural mean squared distance measured between Dino ViT self similarity features as described in \cite{plugnplay}. Lower is better for LPIPS and St.Dist., whereas higher is better for CLIP. The results are provided in Table.\ref{tab:quant}. We can see that different parameters lead to trade off between relevance to edit prompt and structural fidelity to base image. We can identify a set of parameters that gives the best or comparable score relative to the baselines for the proposed technique. We qualitatively observe though that for most edits, the proposed method performs better than all the baselines. To better highlight this disparity, we show edits generated by baselines and PRedItOR on an example base image and edit prompt in Fig.\ref{fig:appendix_unfair} with the corresponding quantitative scores for these edits. We can see that PRedItOR gives the best result but has worser structural fidelity scores than P2P and SD. This is because, the baselines are conservative with the edits retaining the base image almost entirely. Similarly, P\&P gets the best CLIP score but it's corresponding edit has lost all information about the base image. The proposed method does not get the best score for any metric, in this example, while visually does the best when compared with the baselines.

\begin{figure*}[t]
 \centering
 \begin{tabular}{c@{\hspace{0.15cm}}|c@{\hspace{0.15cm}}c@{\hspace{0.15cm}}c@{\hspace{0.15cm}}c@{\hspace{0.15cm}}c@{\hspace{0.15cm}}}
 \centering
 &PRedItOR&P2P&P\&P&SD(SDEdit)&SD(rDDIM)\\
 \hline\\
 \emph{cat}&&&&&\\
 \includegraphics[height=1in, width=1in]{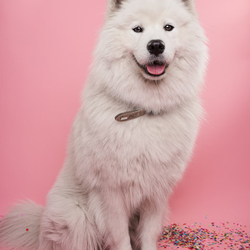}&
 \includegraphics[height=1in, width=1in]{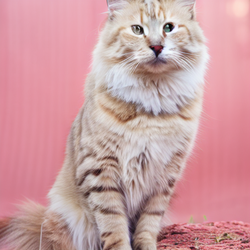}&
 \includegraphics[height=1in, width=1in]{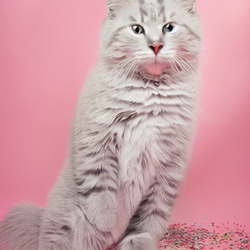}&
 \includegraphics[height=1in, width=1in]{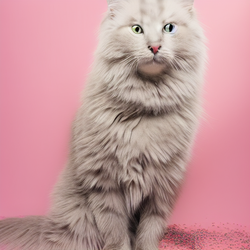}&
 \includegraphics[height=1in, width=1in]{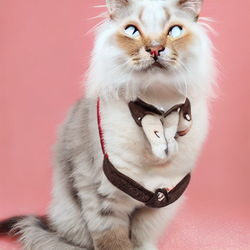}&
 \includegraphics[height=1in, width=1in]{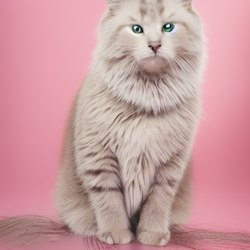} \\ \\
\emph{poodle}&&&&&\\
 \includegraphics[height=1in, width=1in]{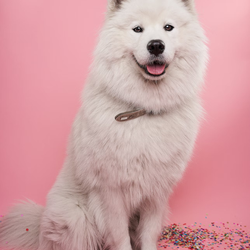}&
 \includegraphics[height=1in, width=1in]{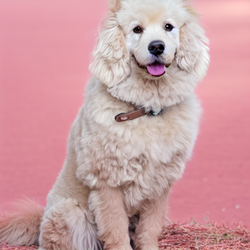}&
 \includegraphics[height=1in, width=1in]{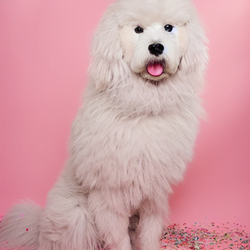}&
 \includegraphics[height=1in, width=1in]{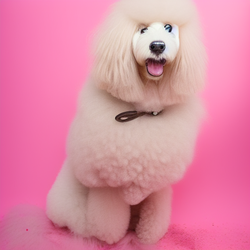}&
 \includegraphics[height=1in, width=1in]{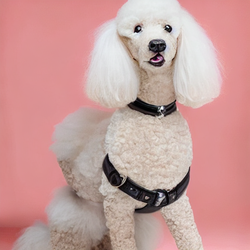}&
 \includegraphics[height=1in, width=1in]{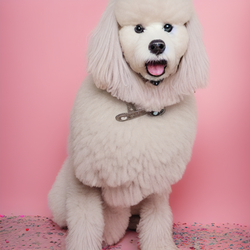} \\ \\
 \emph{oil painting}&&&&&\\
 \includegraphics[height=1in, width=1in]{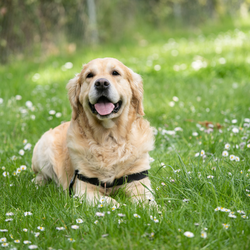}&
 \includegraphics[height=1in, width=1in]{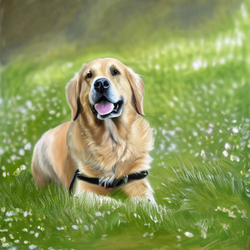}&
 \includegraphics[height=1in, width=1in]{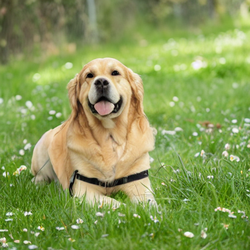}&
 \includegraphics[height=1in, width=1in]{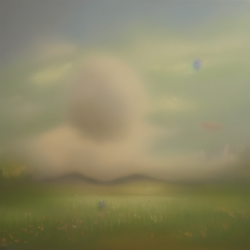}&
 \includegraphics[height=1in, width=1in]{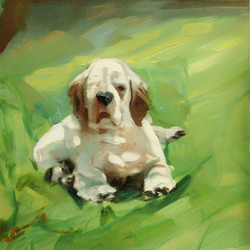}&
 \includegraphics[height=1in, width=1in]{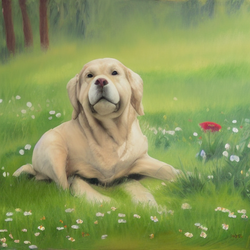} \\ \\
 \emph{children's drawing}&&&&&\\
 \includegraphics[height=1in, width=1in]{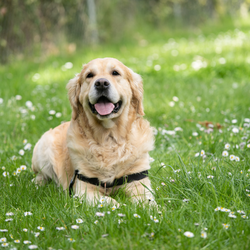}&
 \includegraphics[height=1in, width=1in]{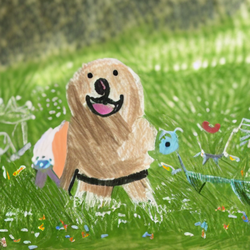}&
 \includegraphics[height=1in, width=1in]{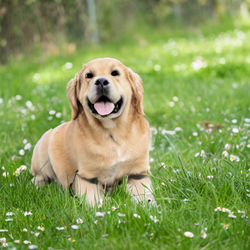}&
 \includegraphics[height=1in, width=1in]{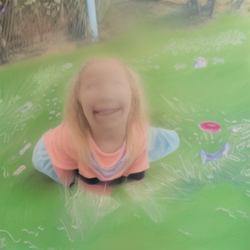}&
 \includegraphics[height=1in, width=1in]{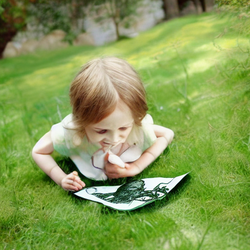}&
 \includegraphics[height=1in, width=1in]{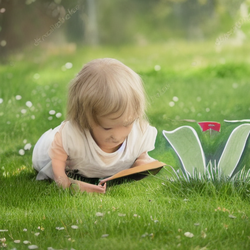} \\ \\
 \emph{stuck in a flood}&&&&&\\
 \includegraphics[height=1in, width=1in]{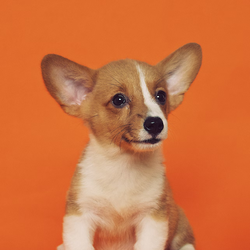}&
 \includegraphics[height=1in, width=1in]{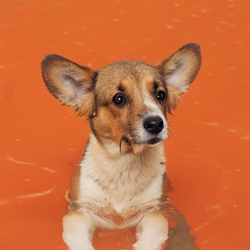}&
 \includegraphics[height=1in, width=1in]{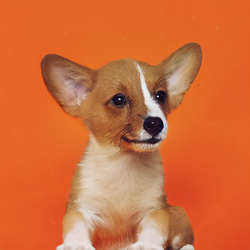}&
 \includegraphics[height=1in, width=1in]{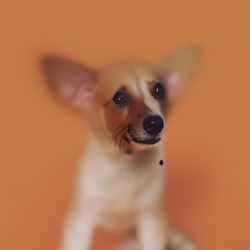}&
 \includegraphics[height=1in, width=1in]{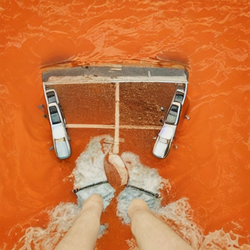}&
 \includegraphics[height=1in, width=1in]{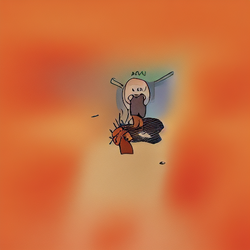}
 \end{tabular}
 \caption{Example edits from the quant eval set from class \emph{dog} comparing PRedItOR with baselines. View the images zoomed in for best results.}
 \label{fig:appendix-quant1}
\end{figure*}

\begin{figure*}[t]
 \centering
 \begin{tabular}{c@{\hspace{0.15cm}}|c@{\hspace{0.15cm}}c@{\hspace{0.15cm}}c@{\hspace{0.15cm}}c@{\hspace{0.15cm}}c@{\hspace{0.15cm}}}
 \centering
 &PRedItOR&P2P&P\&P&SD(SDEdit)&SD(rDDIM)\\
 \hline\\
 \emph{red velvet cake}&&&&&\\
 \includegraphics[height=1in, width=1in]{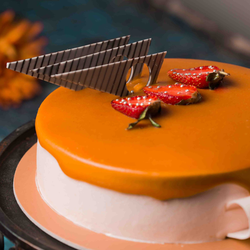}&
 \includegraphics[height=1in, width=1in]{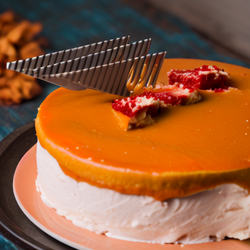}&
 \includegraphics[height=1in, width=1in]{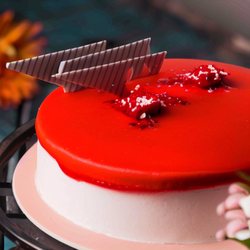}&
 \includegraphics[height=1in, width=1in]{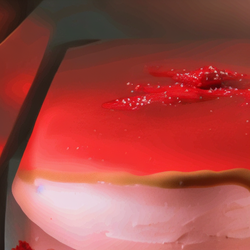}&
 \includegraphics[height=1in, width=1in]{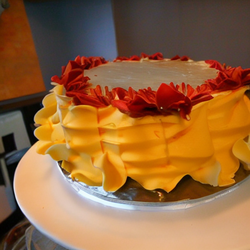}&
 \includegraphics[height=1in, width=1in]{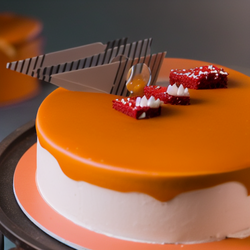} \\ \\
\emph{wedding cake}&&&&&\\
 \includegraphics[height=1in, width=1in]{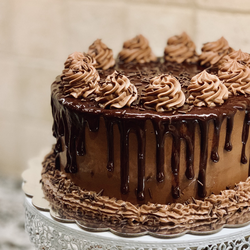}&
 \includegraphics[height=1in, width=1in]{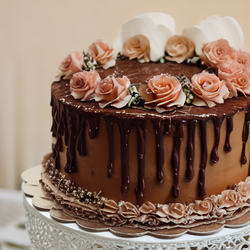}&
 \includegraphics[height=1in, width=1in]{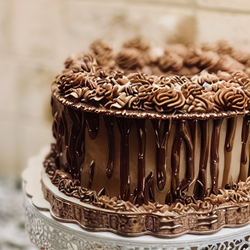}&
 \includegraphics[height=1in, width=1in]{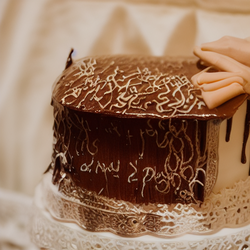}&
 \includegraphics[height=1in, width=1in]{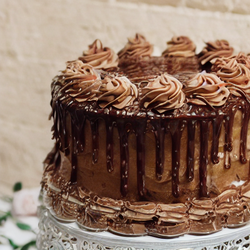}&
 \includegraphics[height=1in, width=1in]{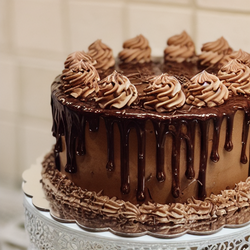} \\ \\
 \emph{oil painting}&&&&&\\
 \includegraphics[height=1in, width=1in]{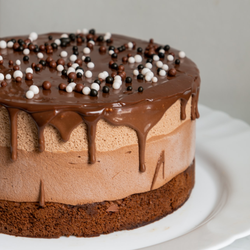}&
 \includegraphics[height=1in, width=1in]{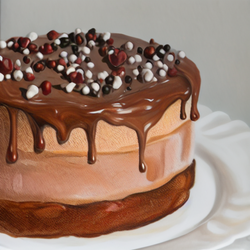}&
 \includegraphics[height=1in, width=1in]{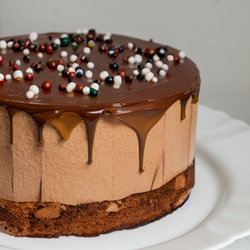}&
 \includegraphics[height=1in, width=1in]{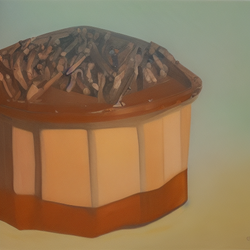}&
 \includegraphics[height=1in, width=1in]{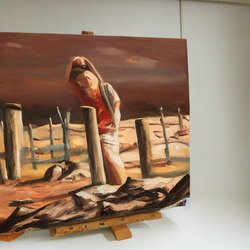}&
 \includegraphics[height=1in, width=1in]{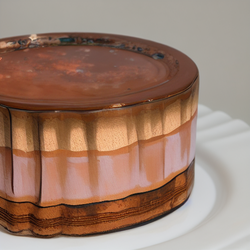} \\ \\
 \emph{children's drawing}&&&&&\\
 \includegraphics[height=1in, width=1in]{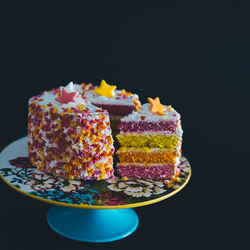}&
 \includegraphics[height=1in, width=1in]{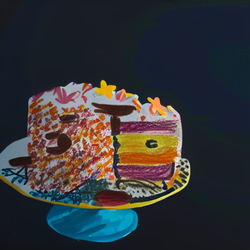}&
 \includegraphics[height=1in, width=1in]{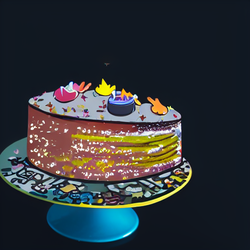}&
 \includegraphics[height=1in, width=1in]{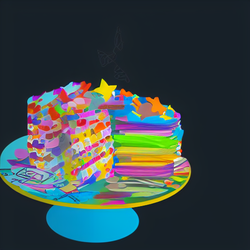}&
 \includegraphics[height=1in, width=1in]{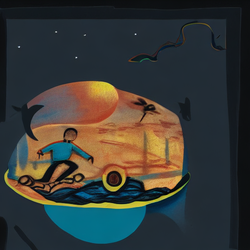}&
 \includegraphics[height=1in, width=1in]{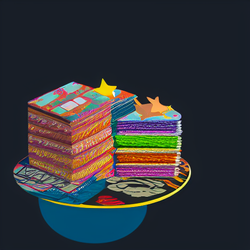} \\ \\
 \emph{lego cake}&&&&&\\
 \includegraphics[height=1in, width=1in]{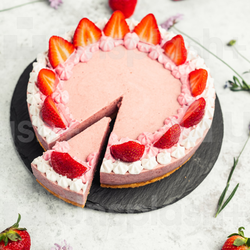}&
 \includegraphics[height=1in, width=1in]{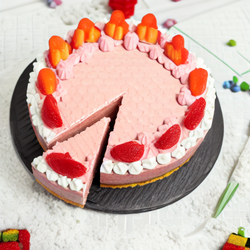}&
 \includegraphics[height=1in, width=1in]{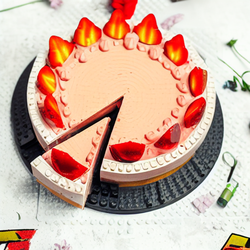}&
 \includegraphics[height=1in, width=1in]{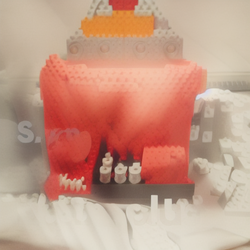}&
 \includegraphics[height=1in, width=1in]{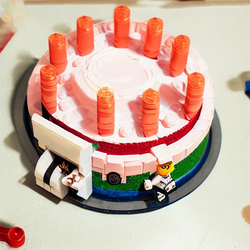}&
 \includegraphics[height=1in, width=1in]{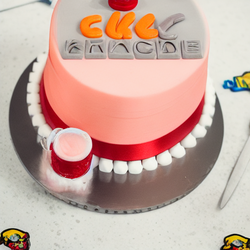}
 \end{tabular}
 \caption{Example edits from the quant eval set from class \emph{cake} comparing PRedItOR with baselines. View the images zoomed in for best results.}
 \label{fig:appendix-quant2}
\end{figure*}

\begin{figure*}[t]
 \centering
 \begin{tabular}{c@{\hspace{0.15cm}}|c@{\hspace{0.15cm}}c@{\hspace{0.15cm}}c@{\hspace{0.15cm}}c@{\hspace{0.15cm}}c@{\hspace{0.15cm}}}
 \centering
 &PRedItOR&P2P&P\&P&SD(SDEdit)&SD(rDDIM)\\
 \hline\\
 \emph{zebra}&&&&&\\
 \includegraphics[height=1in, width=1in]{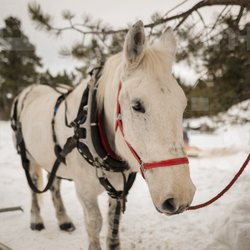}&
 \includegraphics[height=1in, width=1in]{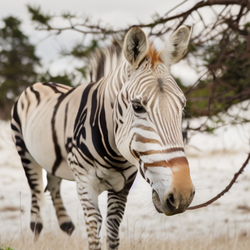}&
 \includegraphics[height=1in, width=1in]{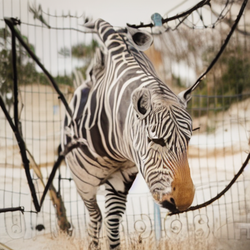}&
 \includegraphics[height=1in, width=1in]{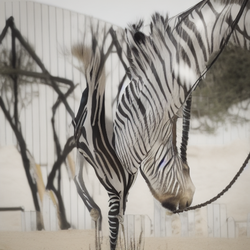}&
 \includegraphics[height=1in, width=1in]{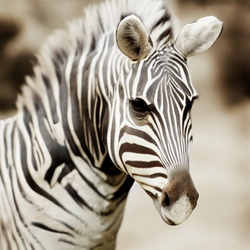}&
 \includegraphics[height=1in, width=1in]{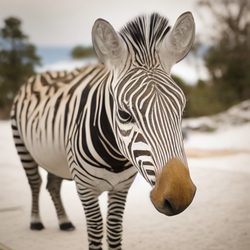} \\ \\
\emph{deer}&&&&&\\
 \includegraphics[height=1in, width=1in]{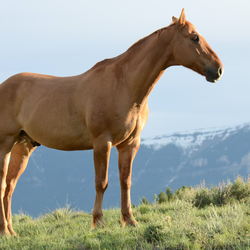}&
 \includegraphics[height=1in, width=1in]{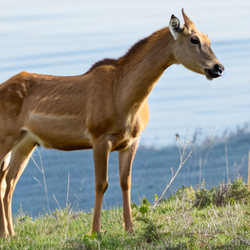}&
 \includegraphics[height=1in, width=1in]{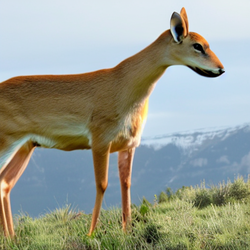}&
 \includegraphics[height=1in, width=1in]{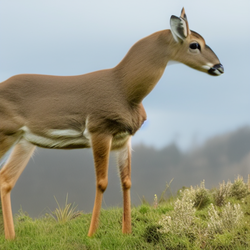}&
 \includegraphics[height=1in, width=1in]{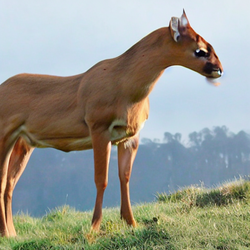}&
 \includegraphics[height=1in, width=1in]{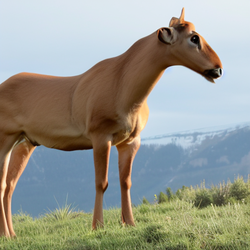} \\ \\
 \emph{oil painting}&&&&&\\
 \includegraphics[height=1in, width=1in]{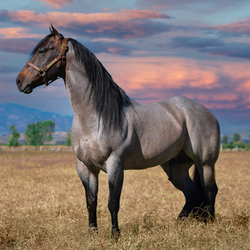}&
 \includegraphics[height=1in, width=1in]{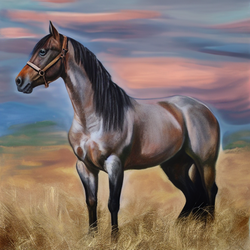}&
 \includegraphics[height=1in, width=1in]{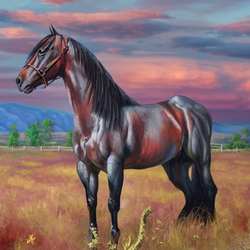}&
 \includegraphics[height=1in, width=1in]{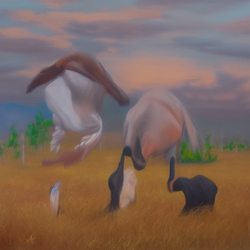}&
 \includegraphics[height=1in, width=1in]{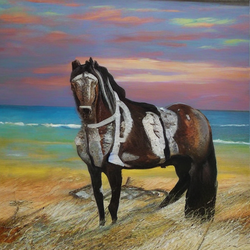}&
 \includegraphics[height=1in, width=1in]{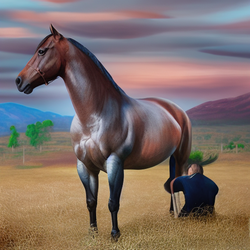} \\ \\
 \emph{children's drawing}&&&&&\\
 \includegraphics[height=1in, width=1in]{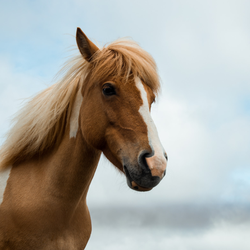}&
 \includegraphics[height=1in, width=1in]{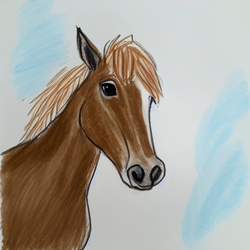}&
 \includegraphics[height=1in, width=1in]{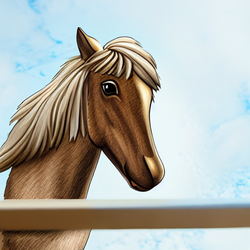}&
 \includegraphics[height=1in, width=1in]{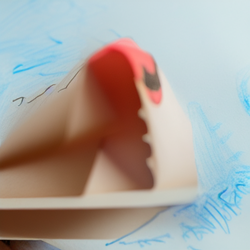}&
 \includegraphics[height=1in, width=1in]{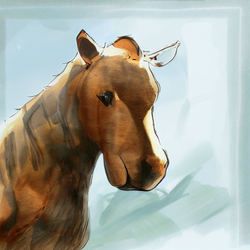}&
 \includegraphics[height=1in, width=1in]{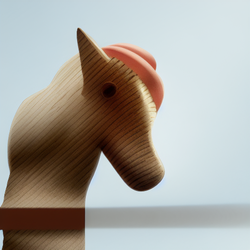} \\ \\
 \emph{stuck in a flood}&&&&&\\
 \includegraphics[height=1in, width=1in]{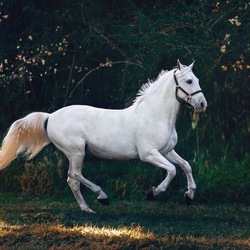}&
 \includegraphics[height=1in, width=1in]{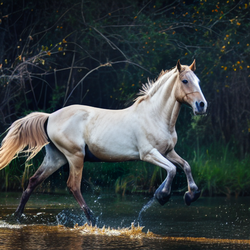}&
 \includegraphics[height=1in, width=1in]{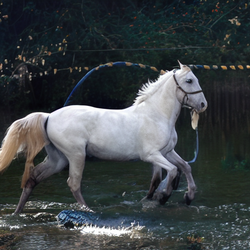}&
 \includegraphics[height=1in, width=1in]{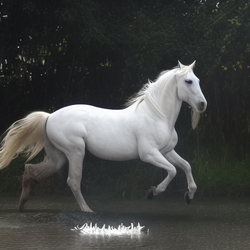}&
 \includegraphics[height=1in, width=1in]{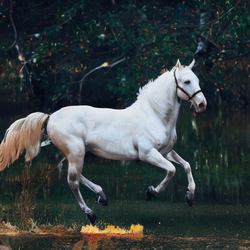}&
 \includegraphics[height=1in, width=1in]{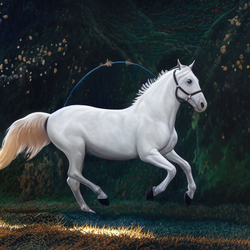}
 \end{tabular}
 \caption{Example edits from the quant eval set from class \emph{horse} comparing PRedItOR with baselines. View the images zoomed in for best results.}
 \label{fig:appendix-quant3}
\end{figure*}

\begin{figure*}[t]
 \centering
 \begin{tabular}{c@{\hspace{0.15cm}}|c@{\hspace{0.15cm}}c@{\hspace{0.15cm}}c@{\hspace{0.15cm}}c@{\hspace{0.15cm}}c@{\hspace{0.15cm}}}
 \centering
 &PRedItOR&P2P&P\&P&SD(SDEdit)&SD(rDDIM)\\
 \hline\\
 \emph{apples}&&&&&\\
 \includegraphics[height=1in, width=1in]{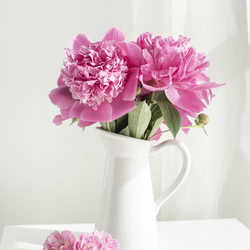}&
 \includegraphics[height=1in, width=1in]{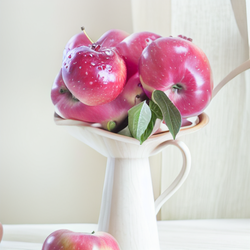}&
 \includegraphics[height=1in, width=1in]{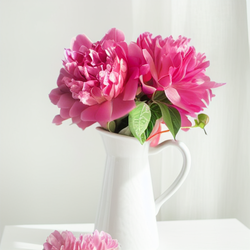}&
 \includegraphics[height=1in, width=1in]{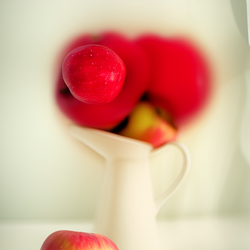}&
 \includegraphics[height=1in, width=1in]{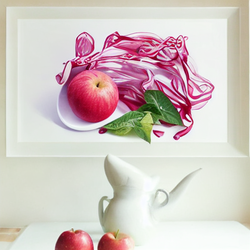}&
 \includegraphics[height=1in, width=1in]{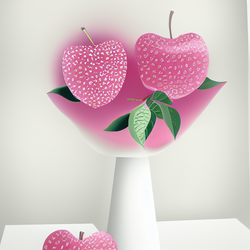} \\ \\
\emph{oranges}&&&&&\\
 \includegraphics[height=1in, width=1in]{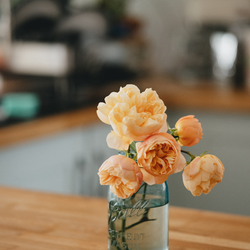}&
 \includegraphics[height=1in, width=1in]{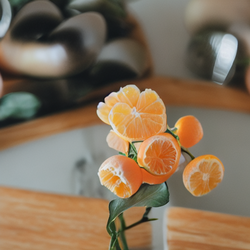}&
 \includegraphics[height=1in, width=1in]{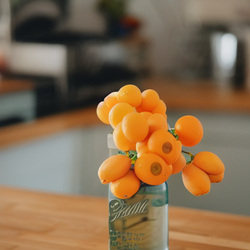}&
 \includegraphics[height=1in, width=1in]{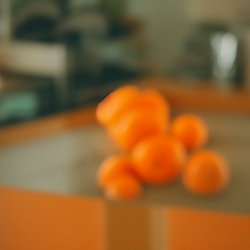}&
 \includegraphics[height=1in, width=1in]{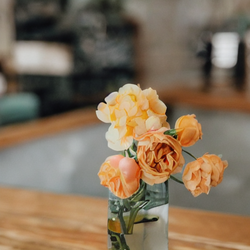}&
 \includegraphics[height=1in, width=1in]{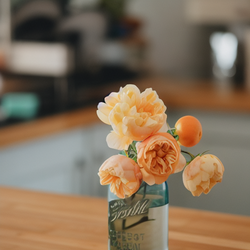} \\ \\
 \emph{oil painting}&&&&&\\
 \includegraphics[height=1in, width=1in]{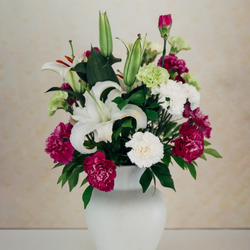}&
 \includegraphics[height=1in, width=1in]{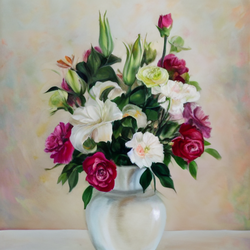}&
 \includegraphics[height=1in, width=1in]{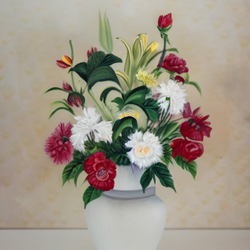}&
 \includegraphics[height=1in, width=1in]{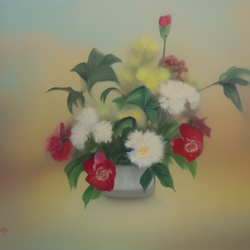}&
 \includegraphics[height=1in, width=1in]{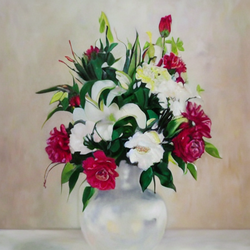}&
 \includegraphics[height=1in, width=1in]{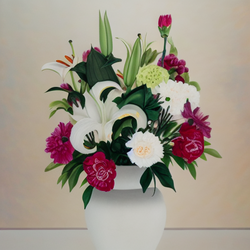} \\ \\
 \emph{children's drawing}&&&&&\\
 \includegraphics[height=1in, width=1in]{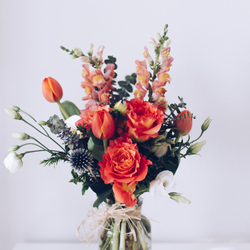}&
 \includegraphics[height=1in, width=1in]{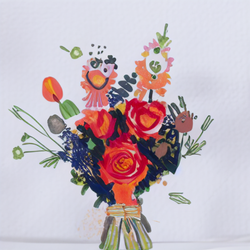}&
 \includegraphics[height=1in, width=1in]{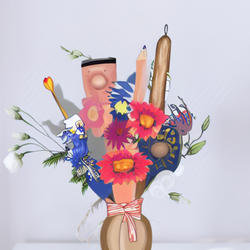}&
 \includegraphics[height=1in, width=1in]{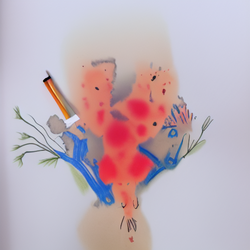}&
 \includegraphics[height=1in, width=1in]{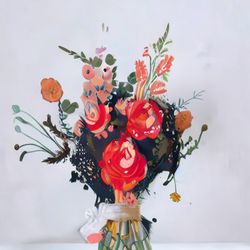}&
 \includegraphics[height=1in, width=1in]{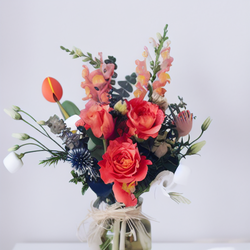} \\ \\
 \emph{origami}&&&&&\\
 \includegraphics[height=1in, width=1in]{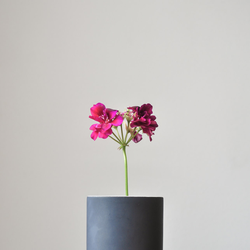}&
 \includegraphics[height=1in, width=1in]{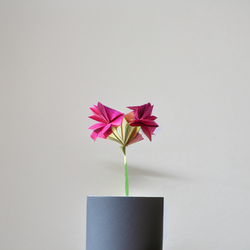}&
 \includegraphics[height=1in, width=1in]{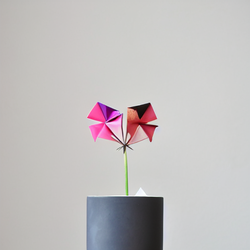}&
 \includegraphics[height=1in, width=1in]{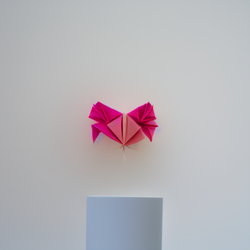}&
 \includegraphics[height=1in, width=1in]{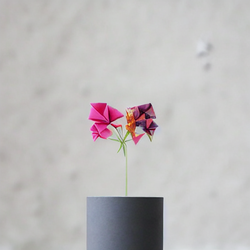}&
 \includegraphics[height=1in, width=1in]{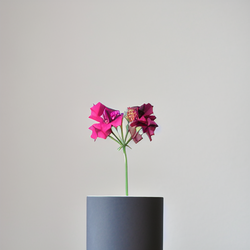}
 \end{tabular}
 \caption{Example edits from the quant eval set from class \emph{flowers} comparing PRedItOR with baselines. View the images zoomed in for best results.}
 \label{fig:appendix-quant4}
\end{figure*}

We provide qualitative example from baselines as well as proposed for all classes of images and edit prompts from the quantitative eval set in Fig.\ref{fig:appendix-quant1}, \ref{fig:appendix-quant2}, \ref{fig:appendix-quant3} and \ref{fig:appendix-quant4} to visually analyze the edits. We can clearly see that for most cases, our proposed method gives the best results visually though that is not directly reflected in the quantitative metrics. P\&P shows subpar generations and we believe some of this can be attributed to the number of sampling steps for inversion being reduced to 50 in our implementation. PRedItOR on the other hand does not require memory intensive or time consuming computations to perform edits. We also highlight that we choose to use P2P with null inversion to improve P2P's results for real images. We believe using P2P with default reverse DDIM based inversion would have generated worser results.  

\section{Additional Examples}
\label{appendix:examples}
\begin{figure*}[t]
 \centering
 \begin{tabular}{c@{\hspace{0.2cm}}|c@{\hspace{0.2cm}}c@{\hspace{0.2cm}}c@{\hspace{0.2cm}}c@{\hspace{0.2cm}}}
 \centering
 & \emph{a mouse}&\emph{a robot}&\emph{pomeranian}&\emph{van gogh style}\\
 \includegraphics[height=1.3in, width=1.3in]{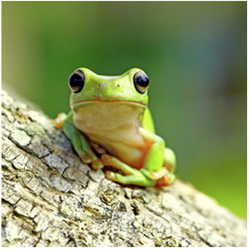}&
 \includegraphics[height=1.3in, width=1.3in]{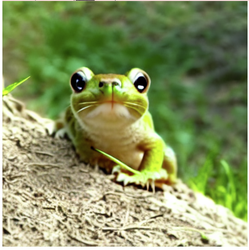}&
 \includegraphics[height=1.3in, width=1.3in]{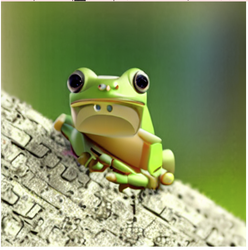}&
 \includegraphics[height=1.3in, width=1.3in]{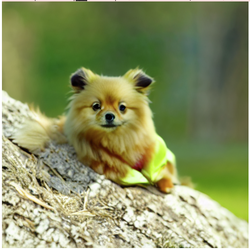}&
 \includegraphics[height=1.3in, width=1.3in]{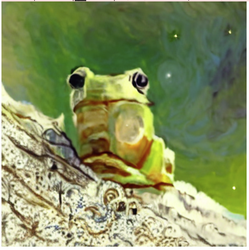} \\ \\
 & \emph{a dog}&\emph{pencil sketch}&\emph{origami}&\emph{a robot}\\
 \includegraphics[height=1.3in, width=1.3in]{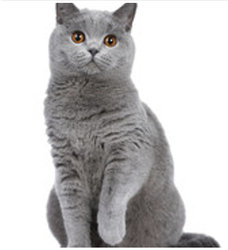}&
 \includegraphics[height=1.3in, width=1.3in]{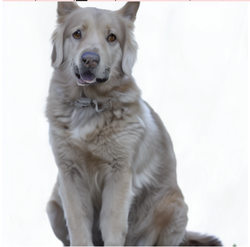}&
 \includegraphics[height=1.3in, width=1.3in]{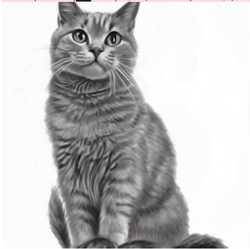}&
 \includegraphics[height=1.3in, width=1.3in]{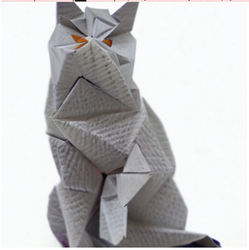}&
 \includegraphics[height=1.3in, width=1.3in]{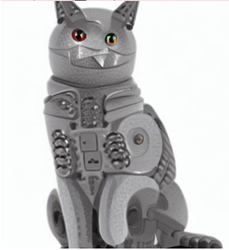} \\ \\
 & \emph{silver cat sculpture}&\emph{sleeping cat}&\emph{tiger}&\emph{watercolor painting}\\
 \includegraphics[height=1.3in, width=1.3in]{images/examples/eg2_real.png}&
 \includegraphics[height=1.3in, width=1.3in]{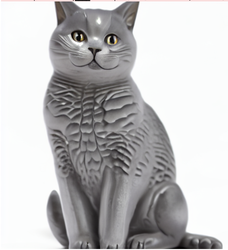}&
 \includegraphics[height=1.3in, width=1.3in]{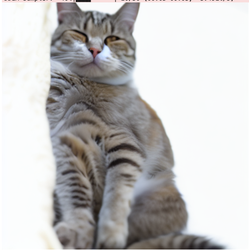}&
 \includegraphics[height=1.3in, width=1.3in]{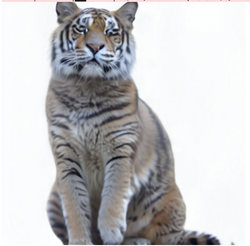}&
 \includegraphics[height=1.3in, width=1.3in]{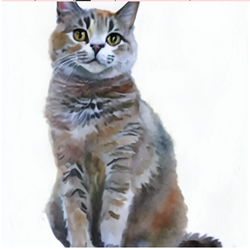} \\ \\
 & \emph{apples}&\emph{planets}&\emph{leaves}&\emph{flowers}\\
 \includegraphics[height=1.3in, width=1.3in]{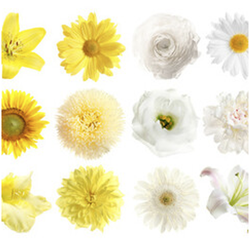}&
 \includegraphics[height=1.3in, width=1.3in]{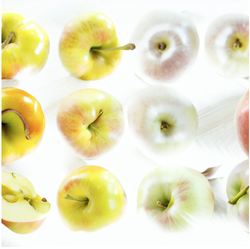}&
 \includegraphics[height=1.3in, width=1.3in]{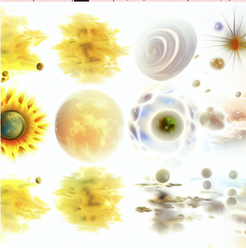}&
 \includegraphics[height=1.3in, width=1.3in]{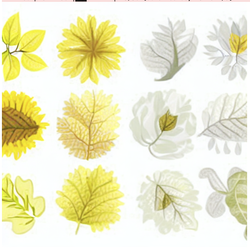}&
 \includegraphics[height=1.3in, width=1.3in]{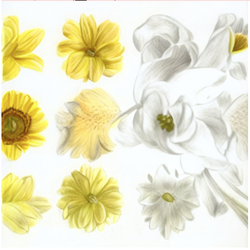} \\ \\
 & \emph{cake with jellybeans}&\emph{pistachio cake}&\emph{red velvet cake}&\emph{chocolate cake}\\
 \includegraphics[height=1.3in, width=1.3in]{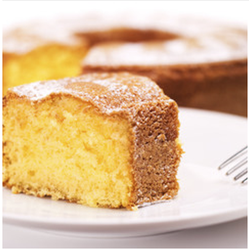}&
 \includegraphics[height=1.3in, width=1.3in]{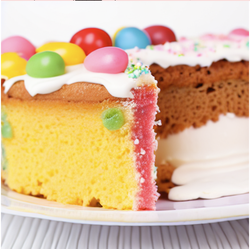}&
 \includegraphics[height=1.3in, width=1.3in]{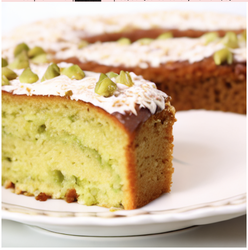}&
 \includegraphics[height=1.3in, width=1.3in]{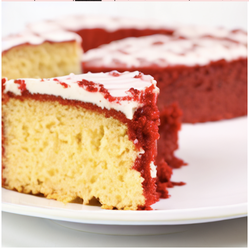}&
 \includegraphics[height=1.3in, width=1.3in]{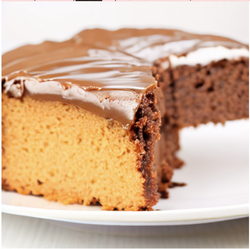} 
 \end{tabular}
 \caption{Example edits covering diverse concepts and scenarios showing the effectiveness of PRedItOR.}
 \label{fig:appendix-examples-real1}
\end{figure*}

\begin{figure*}[t]
 \centering
 \begin{tabular}{c@{\hspace{0.2cm}}|c@{\hspace{0.2cm}}c@{\hspace{0.2cm}}c@{\hspace{0.2cm}}c@{\hspace{0.2cm}}}
 \centering
 & \emph{\parbox{1.3in}{house near a lake with cherry blossoms}}&\emph{... in fall}&\emph{... in spring}&\emph{house on a grass land}\\
 \includegraphics[height=1.3in, width=1.3in]{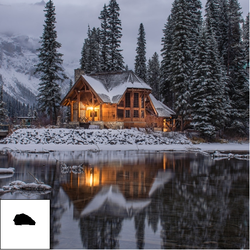}&
 \includegraphics[height=1.3in, width=1.3in]{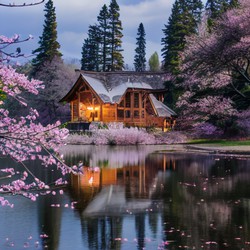}&
 \includegraphics[height=1.3in, width=1.3in]{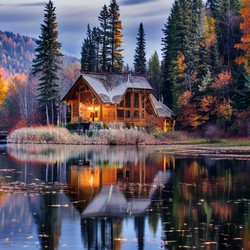}&
 \includegraphics[height=1.3in, width=1.3in]{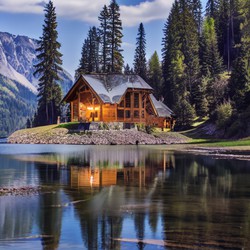}&
 \includegraphics[height=1.3in, width=1.3in]{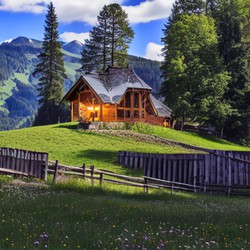} \\ \\
 & \emph{\parbox{1.3in}{childrens drawing of a house near a lake}}&\emph{an oil painting...}&\emph{\parbox{1.3in}{an impressionist painting..}}&\emph{a lowpoly illustration..}\\
 \includegraphics[height=1.3in, width=1.3in]{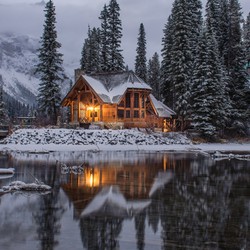}&
 \includegraphics[height=1.3in, width=1.3in]{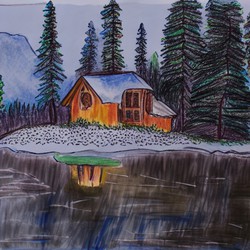}&
 \includegraphics[height=1.3in, width=1.3in]{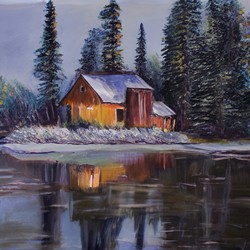}&
 \includegraphics[height=1.3in, width=1.3in]{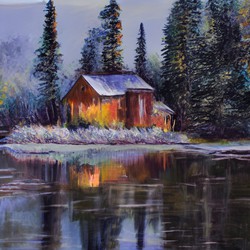}&
 \includegraphics[height=1.3in, width=1.3in]{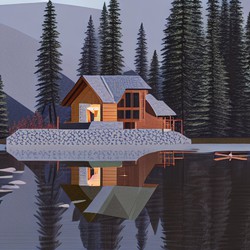} \\ \\
 & \emph{dalmation}&\emph{husky dogs}&\emph{poodle dogs}&\emph{rottweiler}\\
 \includegraphics[height=1.3in, width=1.3in]{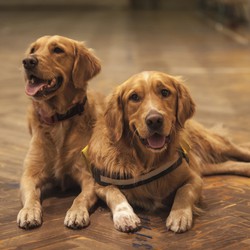}&
 \includegraphics[height=1.3in, width=1.3in]{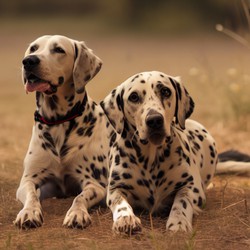}&
 \includegraphics[height=1.3in, width=1.3in]{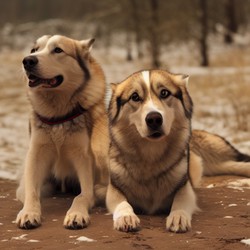}&
 \includegraphics[height=1.3in, width=1.3in]{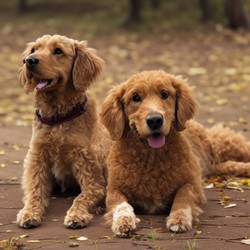}&
 \includegraphics[height=1.3in, width=1.3in]{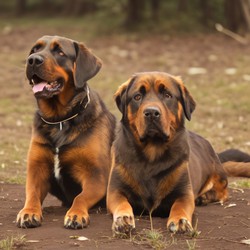} \\ \\
 & \emph{dalmation}&\emph{husky dogs}&\emph{poodle dogs}&\emph{rottweiler}\\
 \includegraphics[height=1.3in, width=1.3in]{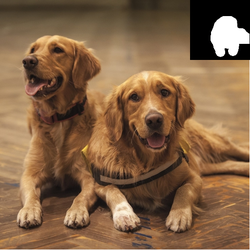}&
 \includegraphics[height=1.3in, width=1.3in]{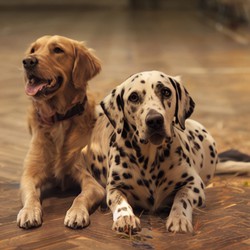}&
 \includegraphics[height=1.3in, width=1.3in]{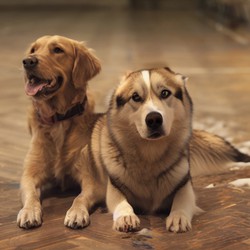}&
 \includegraphics[height=1.3in, width=1.3in]{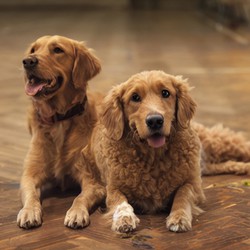}&
 \includegraphics[height=1.3in, width=1.3in]{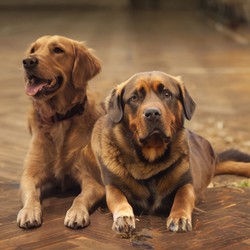} \\ \\
 &\emph{pistachio cake}&\emph{oil painting}&\emph{lego cake}&\emph{wedding cake}\\
 \includegraphics[height=1.3in, width=1.3in]{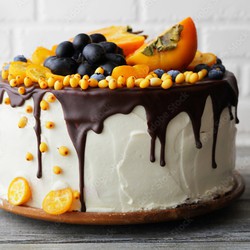} \hspace{0.1in}&
 \includegraphics[height=1.3in, width=1.3in]{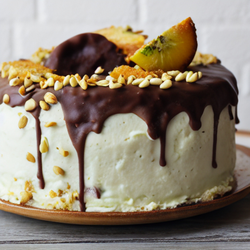}&
 \includegraphics[height=1.3in, width=1.3in]{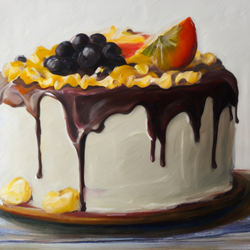}&
 \includegraphics[height=1.3in, width=1.3in]{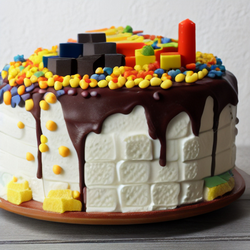}&
 \includegraphics[height=1.3in, width=1.3in]{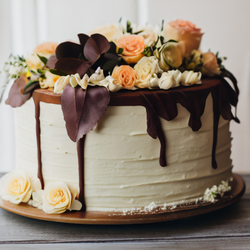} 
 \end{tabular}
 \caption{Example edits showing effective structure preservation by PRedItOR for various edits and the additional benefit when using a mask. Base images are obtained from Unsplash.}
 \label{fig:appendix-examples-real2}
\end{figure*}

\begin{figure*}[t]
 \centering
 \begin{tabular}{c@{\hspace{0.2cm}}|c@{\hspace{0.2cm}}c@{\hspace{0.2cm}}c@{\hspace{0.2cm}}c@{\hspace{0.2cm}}}
 \centering
 & \emph{a camel}&\emph{a cow}&\emph{a giraffe}&\emph{a goat}\\
 \includegraphics[height=1.3in, width=1.3in]{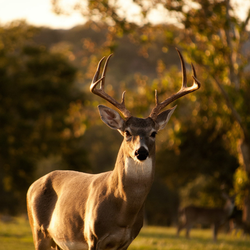}&
 \includegraphics[height=1.3in, width=1.3in]{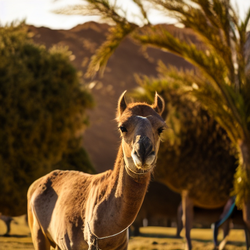}&
 \includegraphics[height=1.3in, width=1.3in]{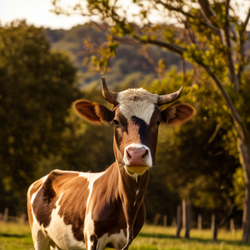}&
 \includegraphics[height=1.3in, width=1.3in]{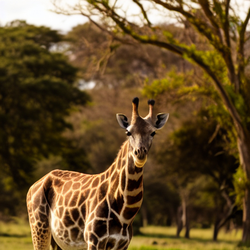}&
 \includegraphics[height=1.3in, width=1.3in]{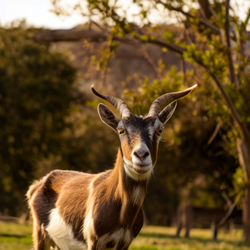} \\ \\
 & \emph{a horse}&\emph{a moose}&\emph{a reindeer}&\emph{a zebra}\\
 \includegraphics[height=1.3in, width=1.3in]{images/examples/appendix_eg1_base.png}&
 \includegraphics[height=1.3in, width=1.3in]{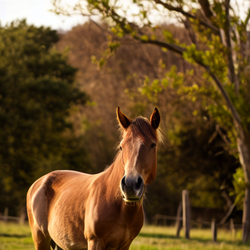}&
 \includegraphics[height=1.3in, width=1.3in]{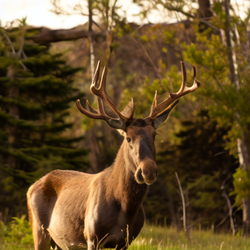}&
 \includegraphics[height=1.3in, width=1.3in]{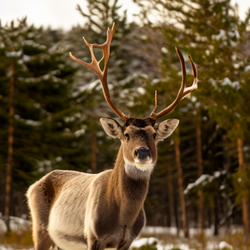}&
 \includegraphics[height=1.3in, width=1.3in]{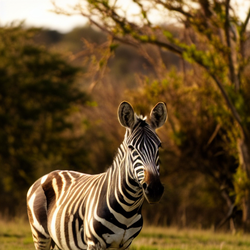} \\ \\
 & \emph{a flying dove}&\emph{... eagle}&\emph{... hummingbird}&\emph{... seagull}\\
 \includegraphics[height=1.3in, width=1.3in]{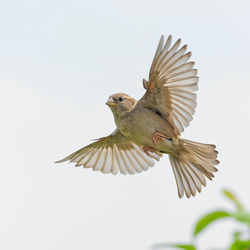}&
 \includegraphics[height=1.3in, width=1.3in]{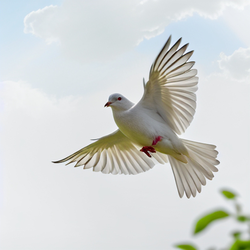}&
 \includegraphics[height=1.3in, width=1.3in]{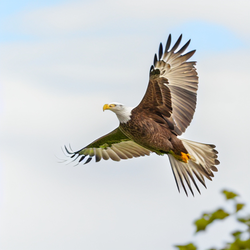}&
 \includegraphics[height=1.3in, width=1.3in]{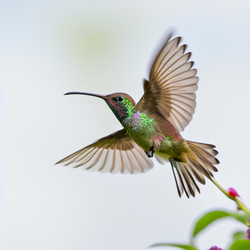}&
 \includegraphics[height=1.3in, width=1.3in]{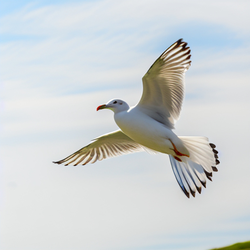} \\ \\
 & \emph{\parbox{1.3in}{lavender field in front of mountains}}&\emph{a tulip field ...}&\emph{a lake ...}&\emph{a sunflower field ...}\\
 \includegraphics[height=1.3in, width=1.3in]{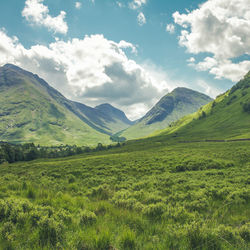}&
 \includegraphics[height=1.3in, width=1.3in]{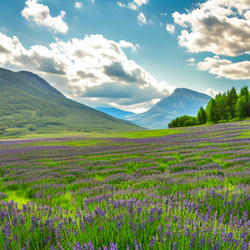}&
 \includegraphics[height=1.3in, width=1.3in]{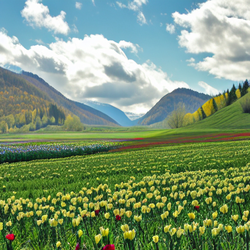}&
 \includegraphics[height=1.3in, width=1.3in]{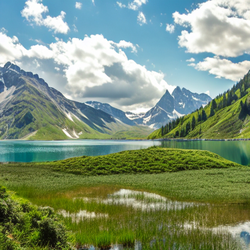}&
 \includegraphics[height=1.3in, width=1.3in]{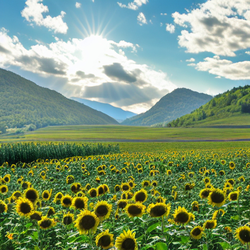} \\ \\
 & \emph{\parbox{1.3in}{a cat wearing sunglasses}}&\emph{a dog}&\emph{a pencil sketch}&\emph{a photorealistic cat}\\
 \includegraphics[height=1.3in, width=1.3in]{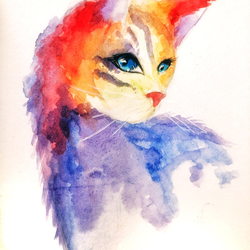}&
 \includegraphics[height=1.3in, width=1.3in]{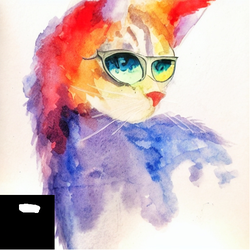}&
 \includegraphics[height=1.3in, width=1.3in]{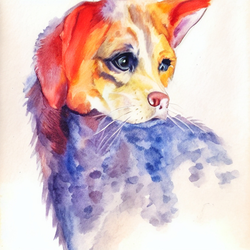}&
 \includegraphics[height=1.3in, width=1.3in]{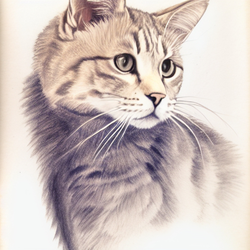}&
 \includegraphics[height=1.3in, width=1.3in]{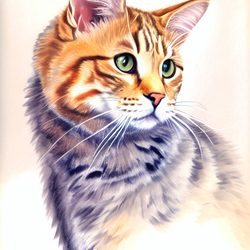} 
 \end{tabular}
 \caption{Example edits showing effective structure preservation by PRedItOR for various edits and the additional benefit when using a mask. Base images are obtained from Unsplash.}
 \label{fig:appendix-examples-real3}
\end{figure*}

We provide more examples of text guided image editing using PRedItOR in Fig.\ \ref{fig:appendix-examples-real1}, Fig.\ \ref{fig:appendix-examples-real2} and Fig.\ \ref{fig:appendix-examples-real3}. These examples were run on images generated by the HDM. We can see from the examples that the proposed PRedItOR can be effectively used to perform text guided editing under a diverse set of scenarios. For example, in Fig.\ \ref{fig:appendix-examples-real1}, we can see changes in objects (frog to mouse, cat to dog) as well as attributes (cat to sleeping) and styles (real photo of a cat to watercolor painting). Moreover, we can further observe that all edits in Fig.\ \ref{fig:appendix-examples-real1} were performed without additional masks or base prompts. 

We can also see the effect of masks in Fig.\ \ref{fig:appendix-examples-real2} in the first and fourth rows. In the first row, the house is unmasked whereas all other regions are masked for the model to edit those regions. As a result, we can perform diverse edits with addition of objects or change of seasons without disturbing the house. In the third row, we can see that PRedItOR effectively performs edits without changing the pose and structure of the dogs. However, there are slight changes in the background such as adding snow for `husky'. In the 4\textsuperscript{th} row, we can see that adding a mask prevents background changes. Moreover it can effectively also change just one of the two dogs.

Examples in Fig.\ \ref{fig:appendix-examples-real3} show complex edits across diverse concepts such as turning a \emph{deer} into a \emph{camel} or \emph{zebra} with the same structure. We can also add attributes like \emph{sunglasses}, though this requires a mask to get satisfactory results.

\section{Limitations}
\label{appendix:limitations}
One of the major limitations of our approach is that the Diffusion Decoder was trained on a significantly smaller set of data compared to Stable Diffusion, DALLE-2 and Imagen, the models that were used by most previous works \cite{nullinversion, imagic, unitune, sdedit, edict, plugnplay, prompt2prompt, shape, instructpix2pix}. Since the data used to train the LDM did not have faces and people, we are unable to show edits comparing previous works on such images.

Another limitation is that the proposed technique relies on reverse DDIM to complete the second \emph{structural edit} step. Both SDEdit \cite{sdedit} and reverse DDIM \cite{ddim, diffae, edict} have been shown to fail for color changing edits. In our examples, whenever color changes happen due to mentions in the prompt (\emph{blue jay} and \emph{northern cardinal} in Fig.\ \ref{fig:intro}), they are subtler and happen only with $s \geq 0.6$. At these values, the background changes are significant and leads to loss in fidelity relative to the base image. Though these artifacts are not as visible in some examples such as in Fig.\ \ref{fig:intro}, for others, it leads to unsatisfactory results. We also believe that any improvements to reverse DDIM or inversion techniques in diffusion models such as \cite{edict, nullinversion} are applicable for the \emph{structural edit} step in our method and can lead to improved results.

As future work, we would like to understand more about the diffusion process in a multimodal embedding space like CLIP. Moreover, even though \emph{conceptual edit} step can provide embeddings that match a given color transforming edit text, the \emph{structural edit} step is constrained to trade-off color with structure. We would like to explore other possibilities for enabling color transforming edits in the proposed setup. We also plan to work on improving the structural preservation in the final edits by leveraging improvements to the reverse DDIM algorithm from existing literature \cite{nullinversion, edict}. 

\clearpage
\clearpage
\end{document}